\documentclass[conference]{IEEEtran}
\IEEEoverridecommandlockouts
\usepackage{amsmath,amssymb,amsfonts}

\makeatletter
\def\ps@headings{%
\def\@oddhead{\mbox{}\scriptsize\rightmark \hfil 
}%
\def\@evenhead{\scriptsize
\hfil \leftmark\mbox{}}%
\def\@oddfoot{}%
\def\@evenfoot{}}
\usepackage{booktabs}
\usepackage{makecell} 
\makeatother 

\usepackage{subcaption}
\usepackage{graphicx}
\usepackage{textcomp}
\usepackage{xcolor}
\def\BibTeX{{\rm B\kern-.05em{\sc i\kern-.025em b}\kern-.08em
    T\kern-.1667em\lower.7ex\hbox{E}\kern-.125emX}}

\usepackage{colortbl}
\usepackage{xcolor}
\usepackage{array}

\usepackage{graphicx}
\usepackage{epstopdf}
\usepackage{cite} 
\usepackage{times}
\usepackage{dsfont} 
\usepackage{cite}
\usepackage{times}
\usepackage{pifont}
\usepackage{multirow}   
\usepackage{tikz}
\usetikzlibrary{shapes.geometric,calc}
   
\usepackage{amsthm}
\usepackage{graphicx}
\usepackage{color}
\usepackage{ifpdf}
\usepackage{epsfig}
\usepackage{latexsym}
\usepackage{amsfonts}
\usepackage{amssymb}
\usepackage{paralist}
\usepackage{comment}
\usepackage{xspace}
\usepackage{mathrsfs}
\usepackage{amssymb}
\usepackage{setspace}
\usepackage{color}
\usepackage{tabularx}

\usepackage{subeqnarray}
\usepackage{amssymb}
\usepackage{url,epsfig,array}
\usepackage{leftidx}
\usepackage{amsmath}
\usepackage[T1]{fontenc}
\usepackage{aecompl}

\usepackage{calligra}
\usepackage{amsmath} 
\usepackage[noend, linesnumbered, ruled]{algorithm2e}

\def\ie{\textit{i.e.}\xspace}

\def\eg{\textit{e.g.}\xspace}

\newcommand{\rotcol}[1]{\rotatebox[origin=c]{20}{\scriptsize\textit{#1}}}
\usepackage[nameinlink,noabbrev]{cleveref} 

\newcommand{\lineref}[1]{Line~\ref{#1}}
\newcommand{\linerefrange}[2]{Lines~\ref{#1}--\ref{#2}}

\makeatletter
\renewcommand{\maketag@@@}[1]{\hbox{\m@th\normalsize\normalfont#1}}%
\makeatother

\begin{document}

\title{\Huge SABlock: Semantic-Aware KV Cache Eviction with Adaptive Compression Block Size}

  \author{ \IEEEauthorblockN{Jinhan Chen, Jianchun Liu, Hongli Xu, Xianjun Gao, Shilong Wang
  }
 \IEEEauthorblockA{
 School of Computer Science and Technology, University of Science and Technology of China\\
 Suzhou Institute for Advanced Research, University of Science and Technology of China\\}
 }


\maketitle

\begin{abstract}
The growing memory footprint of the Key-Value (KV) cache poses a severe scalability bottleneck for long-context Large Language Model (LLM) inference.
While KV cache eviction has emerged as an effective solution by discarding less critical tokens, existing token-, block-, and sentence-level compression methods struggle to balance semantic coherence and memory efficiency.
To this end, we introduce SABlock, a \underline{s}emantic-aware KV cache eviction framework with \underline{a}daptive \underline{block} sizes.
Specifically, SABlock first performs semantic segmentation to align compression boundaries with linguistic structures, then applies segment-guided token scoring to refine token importance estimation.
Finally, for each segment, a budget-driven search strategy adaptively determines the optimal block size that preserves semantic integrity while improving compression efficiency under a given cache budget.
Extensive experiments on long-context benchmarks demonstrate that SABlock consistently outperforms state-of-the-art baselines under the same memory budgets. For instance, on Needle-in-a-Haystack (NIAH), SABlock achieves 99.9\% retrieval accuracy with only 96 KV entries, nearly matching the performance of the full-cache baseline that retains up to 8K entries. Under a fixed cache budget of 1,024, SABlock further reduces peak memory usage by 46.28\% and achieves up to 9.5$\times$ faster decoding on a 128K context length.

\end{abstract}

\begin{IEEEkeywords}
Large language models, Long-context inference, KV cache eviction, Adaptive block size.
\end{IEEEkeywords}

\section{Introduction}
Large Language Models (LLMs)\cite{zhang2022opt, openai2023gpt4, dubey2024llama3} have emerged as indispensable tools for a wide range of downstream tasks, such as document summarization\cite{huang2021efficient} and question answering\cite{kamalloo2023evaluating}. 
LLM inference typically follows an autoregressive paradigm, where each token is generated based on all previously produced tokens. 
During this process, the model computes key and value projections for each token at every attention layer. 
To eliminate redundant computations, modern inference systems employ a Key-Value (KV) cache, which stores these intermediate representations and thereby accelerates decoding \cite{waddington2013kv,ott2019fairseq,yan2025accelerating}.
However, in recent years, as context lengths have increased substantially to capture longer dependencies and support more sophisticated applications, the memory overhead of the KV cache has become a critical bottleneck \cite{yang2024tokenleftbehindreliable}. 
For example, in Llama-3-8B, with a batch size of 16 and a sequence length of 16K, the KV cache consumes about 128 GB of memory, which is more than 8 times larger than the 16 GB memory footprint of model weights. In addition, the size of the KV cache grows linearly with the input sequence length (details in Section 2.1), leading to an increasing memory demand that conflicts with the limited storage capacity available during inference \cite{pope2022scaling}.
This enormous memory footprint not only severely limits the maximum batch size and degrades inference throughput, but also necessitates high-end, expensive hardware, hindering the widespread deployment of long-context LLMs. 
Therefore, the paramount purpose in enabling practical long-context LLM inference is to reduce KV cache memory consumption without sacrificing generation quality.

To address the memory bottleneck, KV cache eviction \cite{streamingllm,h2o,buzz,nacl,snapkv} has been proposed as a promising solution, which compresses the cache by selectively retaining the KV entries of tokens deemed important while discarding the rest. 
Compared with alternatives such as quantization or offloading to external memory, KV eviction directly controls cache size and delivers immediate memory savings during inference. 
The selection of tokens is typically determined by the position-based methods (\eg, sliding windows \cite{streamingllm}) or attention scores that quantify token importance \cite{h2o}.
However, since eviction is inherently irreversible, removing those crucial tokens for future generations will lead to permanent information loss, causing the model to hallucinate or generate incoherent content. 
\begin{figure*}[t]
  \centering
  \includegraphics[width=1.0\textwidth]{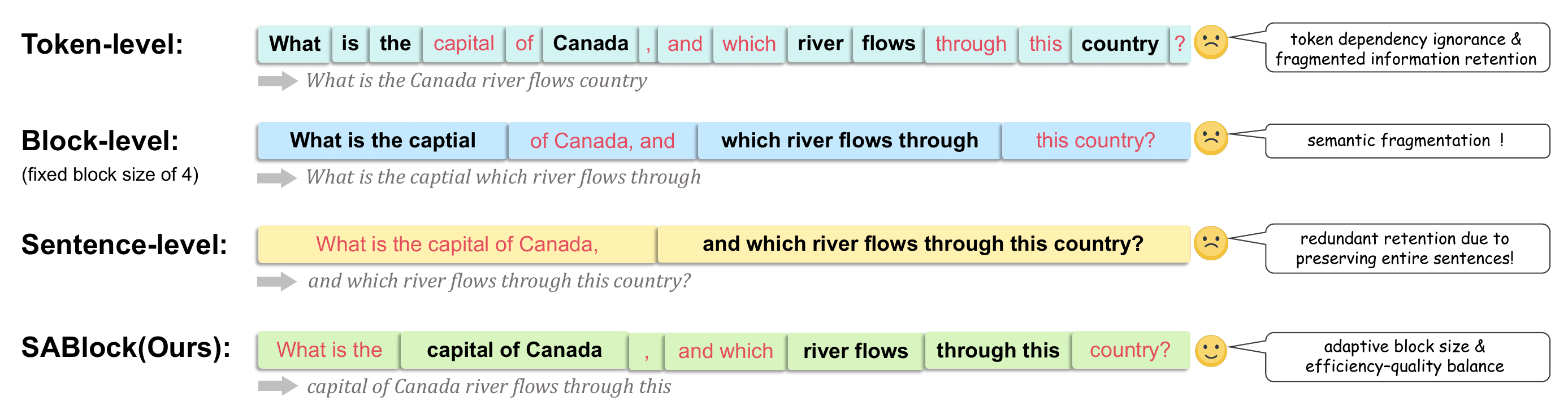}
  \caption{Segmentation strategies applied to the same prompt: token-level, block-level, sentence-level, and our semantic-aware adaptive blocks (the first segment uses a block size of 3 and the second segment uses a block size of 2). 
  \textcolor{red}{Red} tokens indicate evicted content, while \textbf{bold} tokens denote preserved content.}
\label{fig:segmentation_strategies}
\vspace{-5mm}
\end{figure*}
For example, H\textsubscript{2}O\cite{h2o} shows that discarding ``heavy-hitter'' tokens leads to substantial performance drops, such as from about 85\% to 54\% on COPA and from 43\% to 27\% on OpenBookQA, highlighting the risks of erroneous eviction decisions.
Consequently, the challenge for KV cache eviction lies in devising a policy that can \textit{precisely identify and preserve critical tokens, achieving a balance between memory efficiency and generation quality}. 

Among existing KV cache eviction strategies, \textbf{token-level} methods \cite{h2o, snapkv, pyramidkv} represent the most straightforward approach, visualized in Fig.~\ref{fig:segmentation_strategies}. They adopt a greedy selection strategy that progressively retains tokens with higher attention scores, offering fine-grained flexibility but neglecting the semantic dependencies among neighboring words. 
In practice, natural language is not a uniformly distributed sequence of tokens but a hierarchical organization of words, phrases, and sentences. This inherent semantic granularity variation makes compression at the token level prone to disrupting contextual integrity and causing semantic discontinuity or information loss.
Building on this idea, \textbf{block-level} methods \cite{liu2025chunkkv, infllm} group consecutive tokens into fixed-size blocks for compression, aiming to preserve local semantic continuity. 
While this design alleviates fragmentation to some extent, it also introduces new problems. When the block size is too large, a single block may span multiple semantic units and mix unrelated content; when it is too small, the block fails to capture sufficient contextual information. 
In other words, both designs suffer from fundamental semantic limitations. Compression based on a fixed block size struggles to balance semantic coherence and memory efficiency.

From a semantic viewpoint, \textbf{sentence-level} methods \cite{zhu2025sentencekv} further extend this line of work by treating each sentence as an independent compression unit.
Unlike token- or block-level methods that may fragment contextual meaning, this design preserves the complete semantic information within a sentence, thereby maintaining coherence across compression boundaries.
However, sentence lengths vary dramatically, and the distribution of important information across sentences is highly uneven. 
Some sentences convey essential meaning concisely, while others contribute little despite their length.
Under limited cache budgets, retaining entire sentences leads to redundancy and inefficient use of memory resources.

To address these limitations, we propose SABlock, a semantic-aware KV cache eviction framework with adaptive block sizes.
SABlock first performs semantic segmentation to ensure that compression respects natural boundaries.
It then applies a segment-guided token scoring mechanism to propagate segment-level importance to individual tokens, producing more refined scores.
Finally, under a given cache budget, SABlock performs an adaptive block size search for each segment to balance semantic coherence and memory efficiency.
Fig.~\ref{fig:segmentation_strategies} illustrates the segmentation results of different strategies on the same input prompt, demonstrating that our method achieves semantic awareness with adaptive block sizes.

The adaptive block size in SABlock is determined through a budget-driven search strategy.
We observe that the optimal block size depends on the available cache budget: with a tighter budget, smaller blocks better preserve essential details, whereas a relaxed budget favors larger blocks that enable more efficient compression. 
Specifically, SABlock first uses refined token scores to perform global selection, which allocates a local retention budget to each segment. 
Then, within each segment, SABlock performs the search in descending order of candidate block sizes, checking whether the ratio between the importance score at the current block size and that at the token level exceeds a predefined semantic fidelity threshold. 
The largest block size satisfying this condition is selected. 
This procedure preserves coherent semantic structures while adaptively handling non-uniform information density, thus ensuring efficient and reliable cache compression across different budget settings.

Our contributions are summarized as follows:
\begin{itemize}
\item We introduce SABlock, a semantic-aware KV cache eviction framework that jointly enhances generation quality and memory efficiency under varying cache budgets. 
\item We propose a semantic segmentation and segment-guided scoring approach, which ensures compression aligns with natural language boundaries and prevents the loss of critical information.
\item Building on the segmentation and refined token scores, we design a budget-driven adaptive block size search strategy that selects the proper block size satisfying semantic fidelity for each segment.
\item Extensive experiments on long-context benchmarks show that SABlock consistently outperforms state-of-the-art methods across diverse budgets and tasks.
In particular, with a 128K context length, SABlock reduces peak memory usage by 46.28\% and achieves up to 9.5$\times$ faster decoding compared with the full-cache setting.
\end{itemize}

\section{Background and Motivation}
\subsection{KV Cache in LLM Inference}
\begin{figure*}[t]
  \centering



  \newlength{\motH}
  \setlength{\motH}{4.0cm} 

  \begin{subfigure}[t]{0.3\textwidth}
    \centering
    \includegraphics[height=\motH,keepaspectratio]{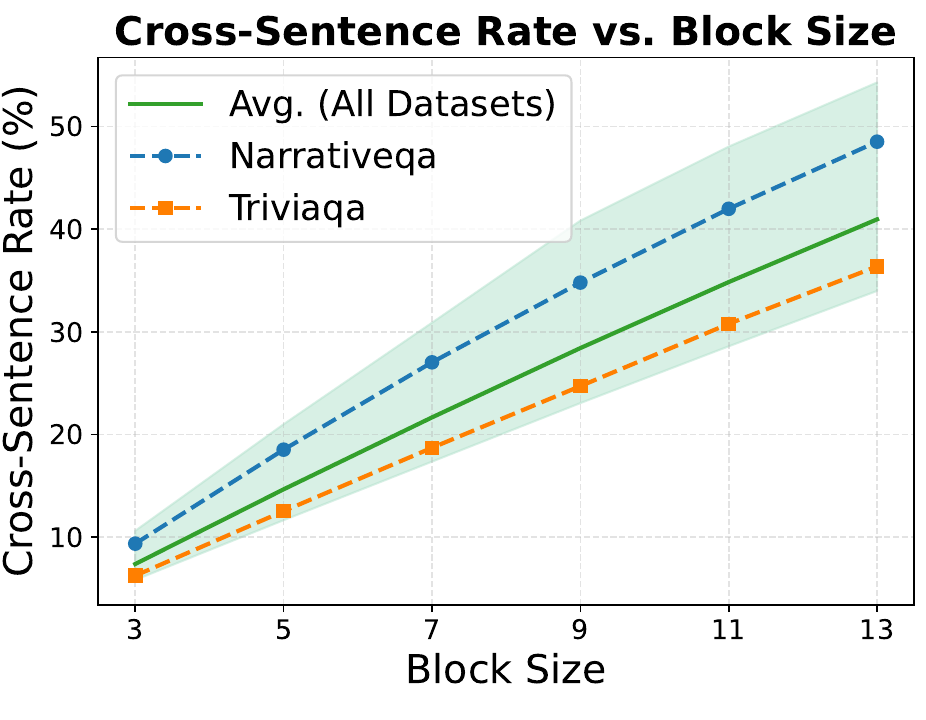}
    \caption{Cross-sentence rate vs. block size.}
    \label{fig:cross_sentence}
  \end{subfigure}
  \hspace{0.025\textwidth}
  \begin{subfigure}[t]{0.3\textwidth}
    \centering
    \includegraphics[height=\motH,keepaspectratio]{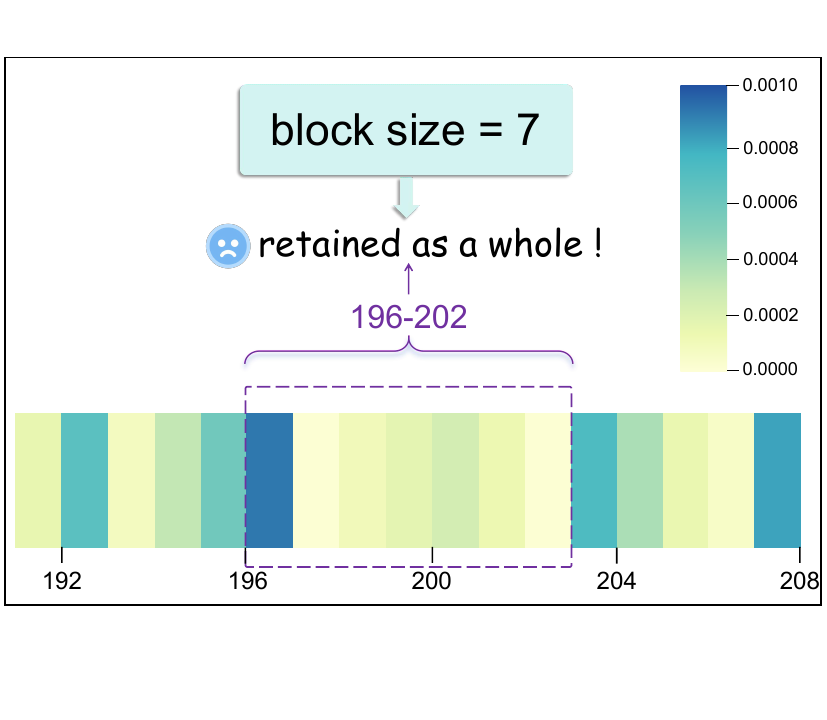}
    \caption{Block-level redundancy illustration.} 
    \label{fig:redundant_retention}
  \end{subfigure}
  \begin{subfigure}[t]{0.3\textwidth}
    \centering
    \includegraphics[height=\motH,keepaspectratio]{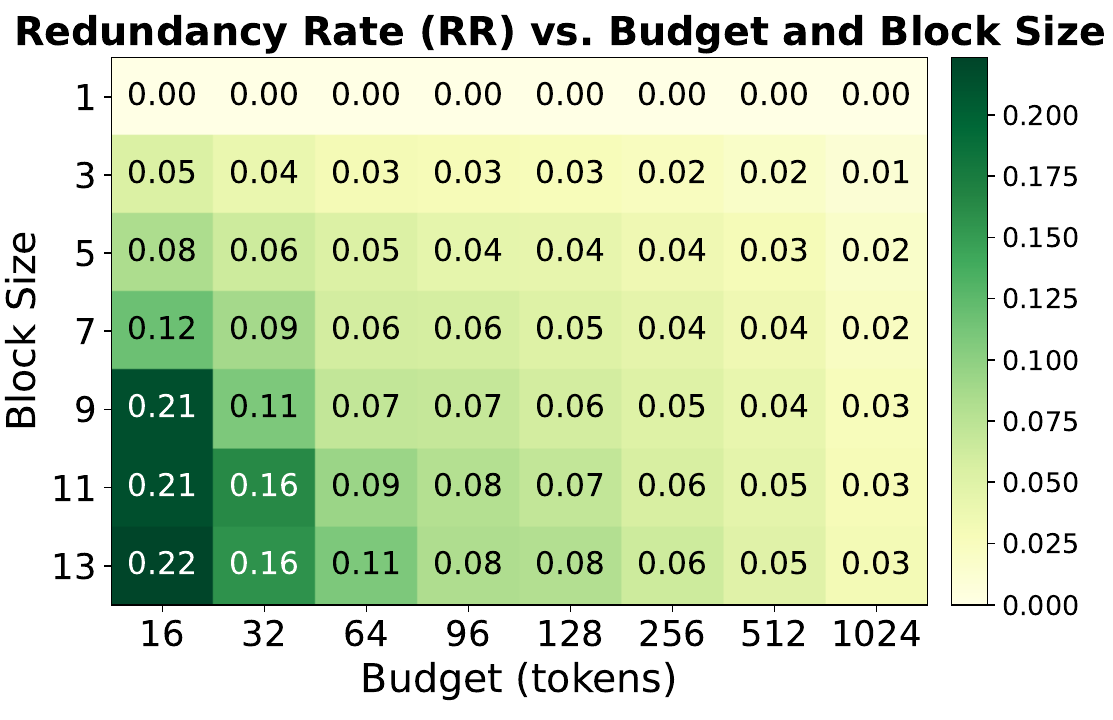}
    \caption{Redundancy rate heatmap.}
    \label{fig:redundant_heatmap}
  \end{subfigure}

  \caption{Illustration of the limitations of block-level KV cache eviction strategies.}
  \label{fig:motivation}
  \vspace{-5mm}
\end{figure*}
LLMs are commonly composed of multiple stacked Transformer decoder layers, where the self-attention mechanism is the core computational component\cite{vaswani2023attentionneed}. 
In this mechanism, the input for the $t$-th token, $x_t \in \mathbb{R}^d$, is transformed into three distinct representations: a query vector ($q_t$), a key vector ($k_t$), and a value vector ($v_t$).
Formally, these vectors are obtained via linear projections of $x_t$:
\begin{equation}
q_t = W_q x_t, \quad k_t = W_k x_t, \quad v_t = W_v x_t
\label{eq:qkv}
\end{equation}
where $W_q$, $W_k$, and $W_v \in \mathbb{R}^{d\times d}$ denote the weight matrices.

Next, the attention output for token $t$ is computed using its query vector $q_t$ and the full history of key and value vectors:
\begin{equation}
\text{Attention}(x_t) = \text{softmax}\left(\frac{q_t K_t^T}{\sqrt{D}}\right) V_t
\label{eq:attention}
\end{equation}
where $K_t = \text{concat}(K_{t-1}, k_t)$, $V_t = \text{concat}(V_{t-1}, v_t)$ (with $K_1 = [k_1]$ and $V_1 = [v_1]$), and $D$ is the key dimension.

Accordingly, the output for the $t$-th token requires all the keys and values from the previous tokens $i \in \{1, \dots, t-1\}$.
To avoid recomputing the representations of previous tokens, these keys and values are stored and reused, forming the KV cache.
During the process, the KV cache is populated and utilized across two distinct phases:
\begin{itemize}
\item Prefill Phase: In this phase, the model first computes the KV pairs for the full input sequence and stores them in the cache for reuse during inference.
\item Decoding Phase: During decoding, the model computes the query, key, and value for the new token, reuses cached keys and values of all previous tokens, and appends the new ones for subsequent steps.
\end{itemize}

The size of the KV cache can be computed as $2 \times b \times l \times L \times h \times c \times p$, where the factor 2 accounts for both keys and values, $b$ is the batch size, $l$ is the number of layers, $L$ is the sequence length, $h$ is the number of attention heads, $c$ is the dimension per head, and $p$ is the precision (\eg, FP16, INT8). As the sequence length grows, the storage demand for the KV cache increases. For example, with a batch size of 32 under FP16, the KV cache requires about 16GB of memory at a sequence length of 32K, and grows to nearly 64GB at 128K. Therefore, the management and optimization of the KV cache are crucial for inference efficiency.

\subsection{Limitations of Existing KV Cache Eviction Methods}

To mitigate the memory overhead of long-context LLM inference, researchers have proposed various KV cache eviction strategies. 
The core idea is to discard less important KV entries to reduce memory usage, while retaining critical information to maintain generation quality.
This creates a fundamental trade-off between memory efficiency and inference performance.
Despite implementation differences, existing methods can be generally categorized by their compression granularity, which defines the basic level at which the KV cache is evaluated and evicted, typically at the token-, block-, or sentence-level.

\textit{1) Token-level eviction methods} evaluate the importance of each token independently and retain only those with the highest attention scores, thereby enabling flexible filtering.
Despite this advantage, these methods still have notable limitations.
Specifically, token-level eviction methods severely disrupt the continuity of semantic information when applied to long contexts. By treating tokens as independent units, these methods disregard the syntactic and semantic dependencies that naturally span across multiple tokens. 
As a result, some tokens that appear unimportant, such as function words or digits in a sequence, may be removed even though they are indispensable parts of larger semantic structures. 
Once such tokens are discarded in isolation, the integrity of the entire segment is broken, and the model can no longer reconstruct the underlying meaning. 
This problem is exacerbated in long-context scenarios, where semantic dependencies are highly entangled and extend over long spans. 
In such cases, token-level eviction not only leads to fragmented context but also introduces severe information loss, making the generated outputs prone to errors, omissions, or hallucinations. 

This limitation has been substantiated by prior work. 
For instance, SnapKV \cite{snapkv} observes that simply selecting tokens with the highest attention scores often undermines information integrity, leading to cases where an LLM may retain only partial details, such as the country code of a phone number, while hallucinating the rest. 
The authors attribute this issue to the tendency of attention mechanisms to assign disproportionately high weights to the initial tokens within a cluster, which makes naive token-level compression particularly harmful. 
Their ablation studies further confirm that token-level methods without additional mechanisms (\eg, pooling) suffer from significant retrieval accuracy degradation, thereby providing concrete evidence of the inherent weakness of token-level eviction.
Thus, while token-level eviction methods offer flexible information filtering, their performance degrades significantly in tasks requiring semantic continuity and context integrity, especially over long texts and in complex scenarios.

\textit{2) Block-level eviction methods} outperform token-level methods in preserving semantic coherence, but they still face two significant limitations that undermine their effectiveness.

\textbf{First, block-level methods are prone to semantic fragmentation due to their reliance on a fixed block size.} 
This approach creates a fundamental mismatch, as the rigid, uniform length of blocks is inherently misaligned with the variable, unbounded length of natural language's semantic units, such as sentences or phrases. 
Consequently, a single block may span across multiple semantic units, causing the information within the block to lose coherence, which in turn affects semantic understanding and reasoning \cite{zhu2025sentencekv}. 
This disregard for natural linguistic boundaries is known to degrade generation performance \cite{transformer-xl, zhu2025sentencekv}.
Previous research indicates that overlooking natural boundaries, like sentence boundaries, can negatively affect language model performance\cite{transformer-xl, zhu2025sentencekv}. 
For example, a block might simultaneously contain the subject of one sentence and the object of another, leading to grammatical and semantic confusion, thereby impacting the accuracy of the generated results. 
To validate this issue, we conduct an analysis on the LongBench benchmark \cite{bai2023longbench}. 
As illustrated in Fig.~\ref{fig:motivation}(a), our results show that even a modest block size of 7 causes an average of 21.65\% of blocks crossing multiple semantic segments, which leads to semantic fragmentation. Moreover, as the block size increases, the likelihood of a block crossing semantic boundaries also rises, exacerbating the issue of semantic fragmentation.
\begin{figure}[t]
\centering
\includegraphics[width=0.88\linewidth]{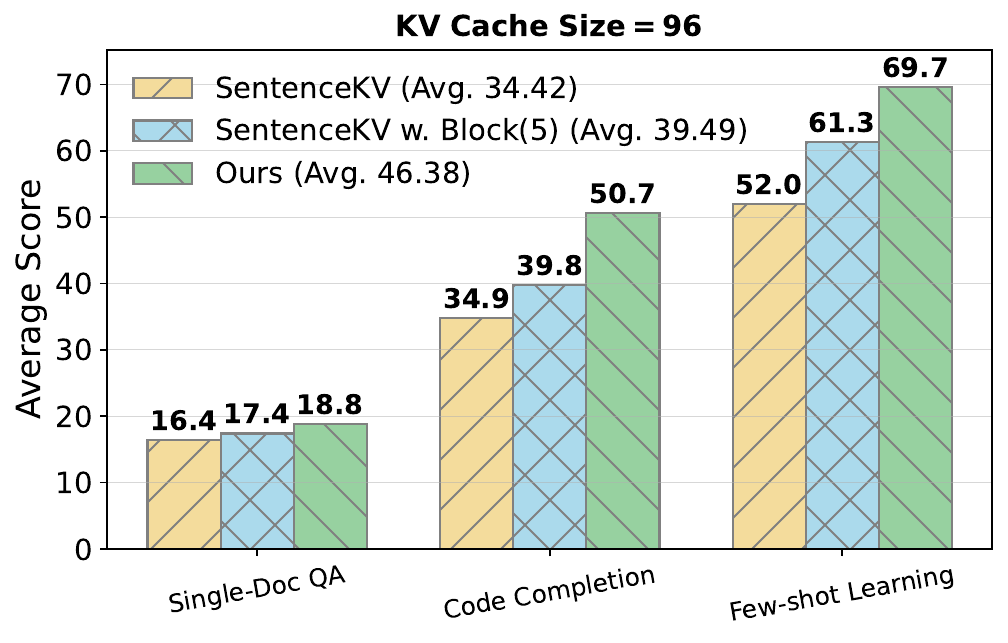}
\caption{Performance comparison under a fixed KV cache budget of 96 on three categories of the LongBench benchmark. We compare SentenceKV, SentenceKV with an additional block size of 5 (SentenceKV w. Block(5)), and our method.}
\label{fig:longbench_kv96}
\vspace{-5mm}
\end{figure}

\textbf{Second, block-level methods often retain many uninformative tokens, leading to inefficient utilization of memory resources.} 
Block-level eviction methods typically require that an entire block be either fully retained or evicted, creating an unavoidable trade-off. When a block contains a mix of important and unimportant tokens, the system is forced to preserve the entire chunk just to save a few critical pieces of information. This policy forces the retention of a large number of uninformative tokens, leading to significant memory waste.
As shown in Fig.~\ref{fig:motivation}(b), given the block size of 7, tokens 196–202 are processed together. Although token 196 has a high attention score, retaining it also forces the retention of low-attention tokens 197–202, which contribute little to semantic coherence. 
To quantify this effect, we measure the redundancy rate, defined as the fraction of attention scores wasted on non-essential tokens relative to a token-level baseline.
As our results in Fig.~\ref{fig:motivation}(c) show, this redundancy rate grows substantially with the block size, confirming the approach's inherent inefficiency.
\begin{figure*}[t]
    \centering
    \includegraphics[width=1.0\textwidth]{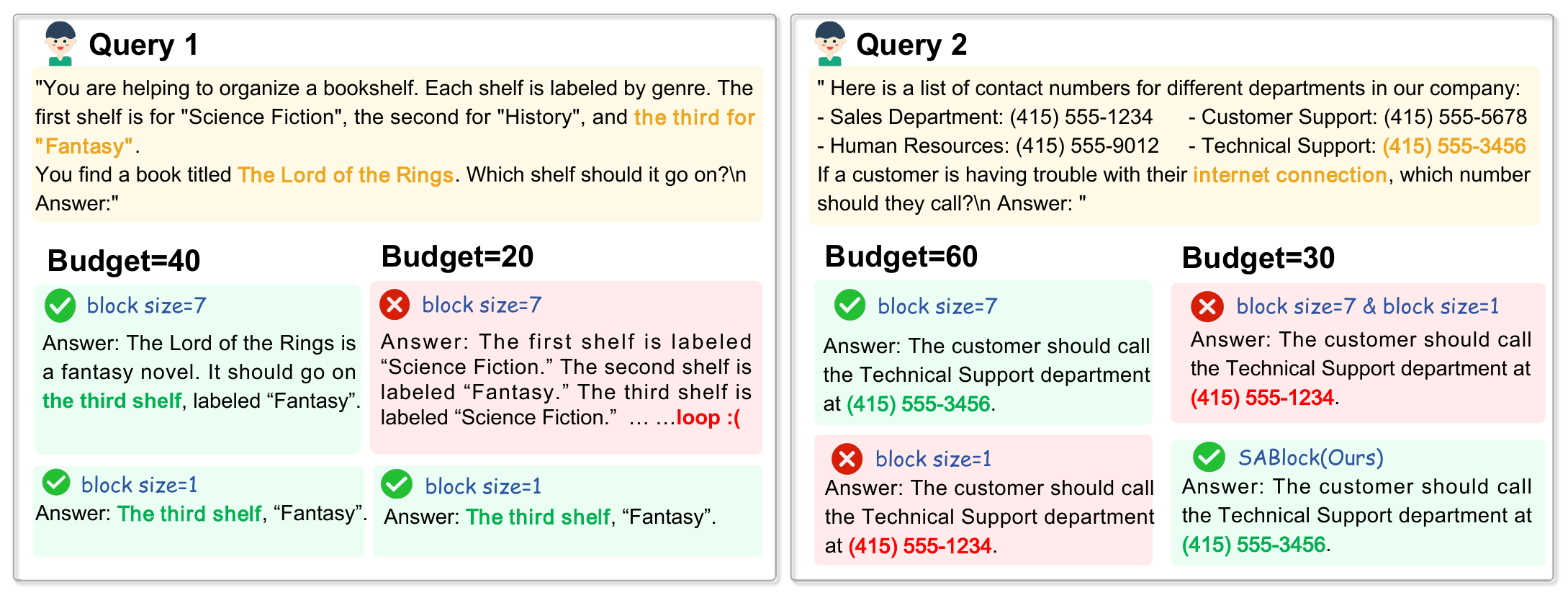} 
    \caption{Answer accuracy under different cache budgets and compression block sizes. The first example demonstrates that compression with a small block size better retains critical information under a low budget, while the second shows that our adaptive method succeeds when fixed block sizes fail.}
    \label{fig:qa_cases} 
\end{figure*}

\textit{3) Sentence-level eviction methods}\cite{zhu2025sentencekv} further improve upon block-level compression by extending the basic compression unit from fixed-length blocks to natural-language sentences. 
This design enables the cache structure to follow the semantic boundaries of sentences, thereby preserving complete semantic information. By retaining sentences as integral units, sentence-level methods maintain semantic integrity and contextual coherence to a large extent. 
However, despite these advantages, these methods still face fundamental limitations that restrict their efficiency. 
Specifically, sentence lengths vary dramatically, and the distribution of information within sentences is highly uneven, \ie, non-uniform information density. 
Some sentences condense crucial content within a few tokens, while others are long but semantically sparse. Under strict cache budgets, retaining entire sentences inevitably keeps a large number of uninformative tokens, leading to severe redundancy and inefficient utilization of memory resources. 
As a result, sentence-level eviction methods tend to preserve a significant amount of low-value content. 

To reduce redundancy within sentences, one possible improvement is to partition each sentence into fixed-size sub-blocks, thereby enabling finer-grained selection. However, this approach still lacks flexibility and cannot adapt to the actual distribution of information. As shown in Fig.~\ref{fig:longbench_kv96}, such partitioning provides only limited performance gains. For instance, Few-shot Learning improves from 52.0 to 61.3, but the score remains below the 69.7 achieved by our method. These results highlight the need for an adaptive block size mechanism for efficient KV cache compression.

\subsection{Opportunity for Adaptive Block Size Configuration}\
The above analysis reveals that the existing KV cache eviction methods suffer from fundamental limitations. 
Existing token-, block-, and sentence-level eviction methods either disrupt semantic continuity, misalign with semantic units, or redundantly retain irrelevant content.
These issues become more pronounced in long-context scenarios, where semantic dependencies are complex and information density is highly uneven. 
Consequently, the core challenge in efficient KV cache eviction is that no single and fixed block size configuration is universally effective under varying cache budgets.

In practice, even under the same cache budget, the most suitable block size may vary across different segments, since the distribution of information along the sequence is highly uneven. Some regions contain dense and critical contextual information, whereas others consist of relatively sparse or redundant content.
Moreover, the overall cache budget is not fixed, as it varies with factors such as context length, batch size, and hardware environment. When the budget changes, the block size configuration must be adjusted accordingly: larger blocks are preferred when memory is sufficient, while smaller blocks are required under tighter budgets.
This suggests that compression should not rely on a single and fixed block size but instead adopt an adaptive configuration to leverage the complementary strengths of varied block sizes.
\begin{table}[t]
    \renewcommand{\arraystretch}{1.2}
    \centering
    \caption{Average score performance of different block sizes under varying budgets.}
    \label{tab:block_performance}
    \resizebox{\linewidth}{!}{%
    \begin{tabular}{c|ccccc|c}
        \hline
        \rowcolor{gray!20}
         & \multicolumn{5}{c|}{\textbf{Block Size}} &  \\
        \cline{2-6}
        \rowcolor{gray!20}
        \multirow{-2}{*}{\cellcolor{gray!20}\textbf{Budget}} 
         & \textbf{1} & \textbf{3} & \textbf{5} & \textbf{7} & \textbf{9} & \multirow{-2}{*}{\cellcolor{gray!20}\textbf{\makecell{Optimal \\ Block Size}}} \\
        \hline
        16   & \cellcolor{yellow!20}\textbf{22.74} & 22.71 & 22.52 & 22.37 & 22.27 & 1 \\
        32   & 23.35 & \cellcolor{yellow!20}\textbf{24.19} & 24.17 & 24.10 & 24.15 & 3 \\
        64   & 24.15 & 25.44 & \cellcolor{yellow!20}\textbf{25.93} & 25.83 & 25.88 & 5 \\
        96   & 24.47 & 25.69 & 26.66 & \cellcolor{yellow!20}\textbf{27.04} & 27.03 & 7 \\
        128  & 24.80 & 26.08 & 27.01 & 27.57 & \cellcolor{yellow!20}\textbf{27.98} & 9 \\
        256  & 25.48 & 26.96 & 28.01 & \cellcolor{yellow!20}\textbf{28.44} & 28.40 & 7 \\
        512  & 26.94 & 27.73 & 29.06 & 29.18 & \cellcolor{yellow!20}\textbf{29.21} & 9 \\
        1024 & 27.98 & 28.62 & 29.82 & 29.80 & \cellcolor{yellow!20}\textbf{29.95} & 9 \\
        2048 & 30.03 & 29.95 & 30.55 & 30.68 & \cellcolor{yellow!20}\textbf{30.79} & 9 \\
        4096 & 30.92 & 30.73 & 31.02 & 31.00 & \cellcolor{yellow!20}\textbf{31.03} & 9 \\
        \hline
    \end{tabular}
    }
\end{table}

We observe that the choice of compression block size is largely constrained by the available cache budget.
Fig.~\ref{fig:qa_cases} presents two representative question-answering scenarios.
In Query 1, when the cache budget is 40, both a relatively large block size (\eg, 7) and a finer block size (\eg, 1) preserve sufficient context and yield correct answers.
However, when the budget is reduced to 20, the large-block method fails because it must retain entire blocks, leading to the truncation of critical information.
In contrast, the small-block method enables more flexible selection of tokens and maintains correctness.
In Query 2, with a budget of 60, the large-block method better preserves token dependencies and the overall semantic structure.
Meanwhile, the small-block method isolates individual tokens, breaking semantic integrity and producing incorrect answers.
When the budget drops to 30, both fixed-block-size methods fail, while only our proposed adaptive block size method successfully produces the correct answer.
These results demonstrate that a \textit{fixed} block size cannot resolve the trade-off between semantic coherence and memory efficiency across different budgets, highlighting the necessity of adopting adaptive block sizes.

Quantitative experiments further validate this observation. Table~\ref{tab:block_performance} presents the average LongBench benchmark scores obtained with different block sizes under varying budgets, evaluated on LLama-3-8B.
When the budget is extremely limited (\eg, 16), token-level compression (block size = 1) achieves the best score results.
As the budget increases, the optimal block size gradually shifts from 3 to 5, and under sufficient budgets ($\geq$128), the best score performance is achieved with a block size of 9.
However, using large block sizes under small budgets causes severe semantic loss and accuracy degradation, which demonstrates that a fixed block size cannot adapt to different budgets.
Although both case studies and experimental results demonstrate that the compression block size must adapt to the memory budget, incorporating this adaptivity into a coherent selection process remains highly non-trivial.
The core challenge, therefore, is to \textbf{adaptively adjust the block size under a global budget constraint while preserving essential semantic information.}

\section{Framework Design}
\begin{figure}[t]
    \centering
    \includegraphics[width=\linewidth]{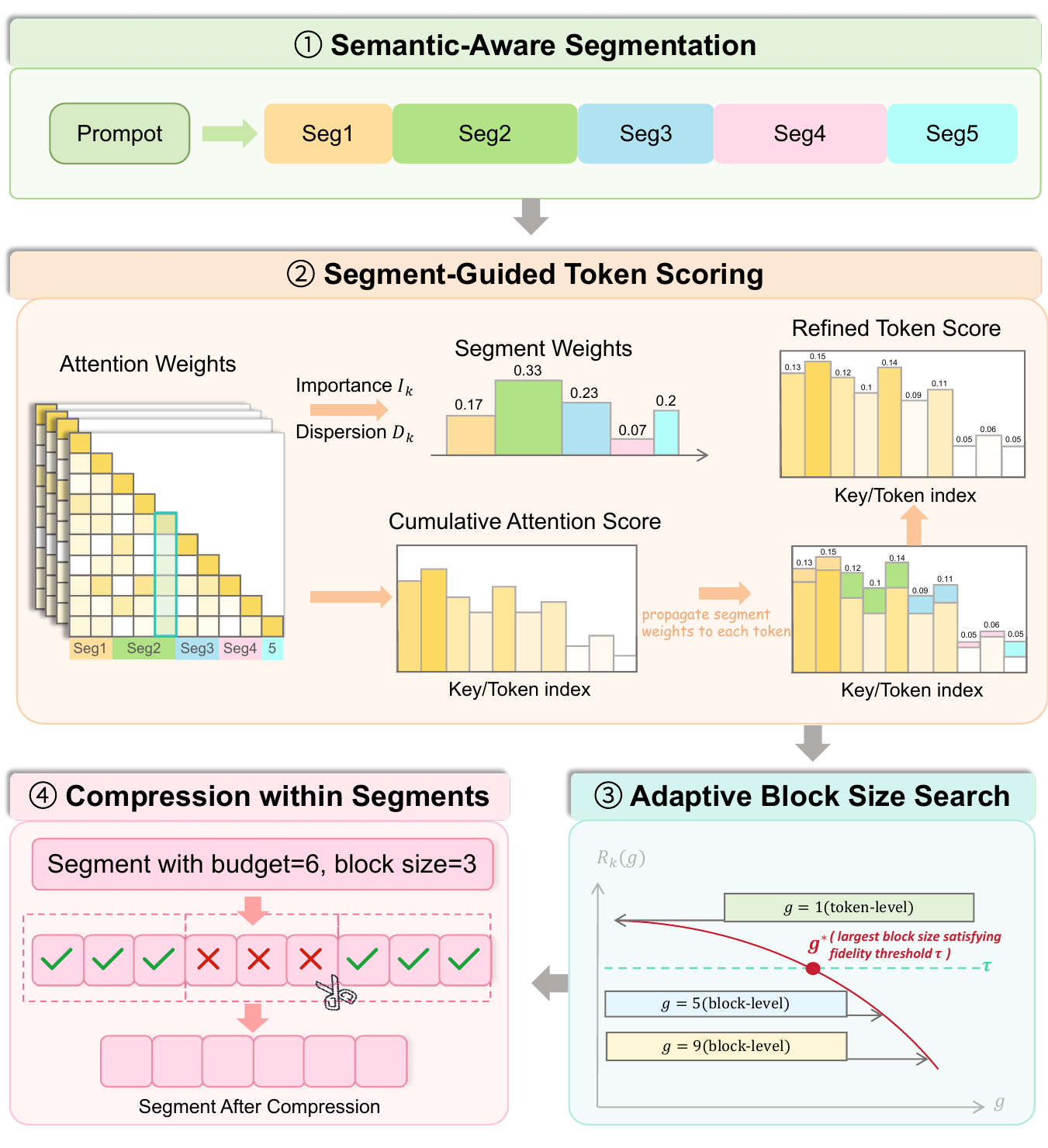} 
    \caption{Overview of SABlock.}
    \label{fig:example} 
    \vspace{-5mm}
\end{figure}

To enable efficient long-context inference under varied memory budgets, we propose SABlock, a semantic-aware KV cache eviction framework with adaptive block sizes.
As illustrated in Fig.~\ref{fig:example}, SABlock ensures semantic awareness through segmentation and segment-guided token scoring, then applies an adaptive block size search within segments to produce the final compressed KV cache. Specifically, SABlock first segments the sequence into semantically coherent spans (\ding{172}) and enhances token-level scoring with segment-level weights to preserve important contextual units (\ding{173}).
On top of this, SABlock introduces a budget-driven adaptive block size search mechanism: given a fixed global cache budget, each segment implicitly receives its share of retained tokens through global selection, and the largest block size that still preserves semantic integrity (\ie, satisfying a predefined fidelity threshold) is then chosen (\ding{174}).
Once the block size is determined, SABlock compresses each segment using the chosen block size and merges the selected tokens to construct the final KV cache (\ding{175}).
This design naturally aligns compression block size with cache budget, thereby achieving a better trade-off between memory efficiency and inference quality.

\subsection{Semantic-Aware Segmentation}\label{sec:segmentation}

Traditional block-level eviction methods typically divide input sequences into fixed-length chunks without regard for natural language structure, which often disrupts semantic continuity and causes critical contextual information to be lost. To improve the coherence of compressed representations, we first segment the sequence into a set of semantically aligned spans $\{ S_1, S_2, \dots, S_k \}$, with each treated as an independent unit in subsequent compression stages.

Before segmentation, we first identify the region that is eligible for compression. Specifically, SABlock distinguishes between two regions: the compressible region and the uncompressed region. The uncompressed region consists of the final $n_{\text{rcnt}}$ tokens in the prompt, which form an observation window that provides crucial local context for subsequent generation. Prior works such as SnapKV~\cite{h2o} adopt a similar approach, always preserving the KV cache for these recent prompt tokens to avoid degrading generation quality. Following this convention, we retain the KV cache for the most recent tokens $\{x_{t - n_{\text{rcnt}} + 1}, \dots, x_t\}$ and apply eviction only to the earlier portion of the prompt $\{x_1, \dots, x_{t - n_{\text{rcnt}}}\}$.

For the compressible region, SABlock performs semantic-aware segmentation in order to reduce information loss across compression boundaries. Although embedding-based segmentation methods are available, such as those that detect shifts in cosine similarity\cite{langchain}, they tend to introduce significant computational overhead. To address this, we adopt a lightweight punctuation-based strategy that leverages natural language delimiters (such as commas and periods) to identify semantically meaningful boundaries. This method strikes a balance between generation accuracy and runtime efficiency, enabling fast segmentation without the need for additional model inference or large memory buffers. The output is a sequence of segments $\{ S_1, S_2, \dots, S_k \}$, each maintaining semantic coherence and serving as an independent unit for the subsequent compression.

\subsection{Segment-Guided Token Scoring}
After performing segmentation, the basic unit of compression becomes semantic segments $\{ S_1, S_2, \dots, S_k \}$ that are aligned with linguistic boundaries. 
However, if SABlock continues to rely solely on the attention scores of individual tokens for global selection, a critical issue emerges. Some segments that are globally important inevitably include tokens whose individual scores are slightly lower, even though these tokens carry essential semantic connections required for maintaining contextual coherence.
Strict point-wise filtering may remove such tokens, inadvertently fragmenting important segments and breaking semantic continuity.

\begin{algorithm}[t]
\caption{\textsc{Segment-Guided Token Scoring}}
\label{alg:spws}
\DontPrintSemicolon
\SetAlgoLined
\SetKw{KwAnd}{and}
\KwIn{Attention matrix $A\in\mathbb{R}^{H\times n_{\text{rcnt}} \times T}$, segments $\{S_k\}$ in the compressible region, hyperparameters $\alpha,\eta,\varepsilon$}
\KwOut{Segment-guided token scores $\tilde{s}_t$ for all $t$ in the compressible region}
\vspace{0.25em}

\textcolor{purple}{\textbf{\# Head-mean attention scores}} \\
Compute mean attention $\bar A_{q,t}$ across heads via (\ref{eq:barA}) \nllabel{spws:barA} \\
\ForEach{token $t$ in $\cup_k S_k$}{
  Compute window-based attention score $s_t$ via (\ref{eq:st}) \nllabel{spws:s_t}
}
\vspace{0.25em}
\textcolor{purple}{\textbf{\# Segment-level statistics computation}} \\
\ForEach{segment $S_k$}{
  $T_k \leftarrow \{\, t : t \in S_k \,\}$ \tcp*{token indices}
  Compute $I_k$ from $\{s_t: t\in T_k\}$ via (\ref{eq:Ik}) \nllabel{spws:Ik} \\
  Compute $D_k$ from $\{s_t: t\in T_k\}$ via (\ref{eq:Dk}) \nllabel{spws:Dk} \\
  Compute segment weight $\omega_k$ via (\ref{eq:omegak}) \nllabel{spws:omega}
}

\vspace{0.25em}
\textcolor{purple}{\textbf{\# Segment-guided token score adjustment}} \\
\ForEach{$t \in R_{\text{mid}}$}{
  Let $k(t)$ be the index such that $t \in S_{k(t)}$ \\
  Compute adjusted token score $\tilde{s}_t$ via (\ref{eq:st_tilde}) \nllabel{spws:st_tilde}
}
\end{algorithm}

To address this issue, SABlock introduces a segment-guided weighting mechanism on top of window-based attention scores. 
As shown in \Cref{alg:spws}, we first compute the head-mean attention and window-based scores for each token (\linerefrange{spws:barA}{spws:s_t}). 
Next, for each segment $S_k$, two statistics (\ie, the average importance $I_k$ and the entropy-based diversity $D_k$) are calculated. These quantities jointly characterize both the overall importance and the distributional spread of attention within segment $S_k$ (\lineref{spws:omega}).
Finally, each token score is adjusted by a factor derived from the statistics of its corresponding segment (\lineref{spws:st_tilde}). This ensures that globally important segments are not fragmented during selection.

Specifically, let $A\in\mathbb{R}^{H\times n_{\text{rcnt}} \times T}$ denote the attention matrix computed from the observation window\cite{snapkv}, where $H$ is the number of heads, $n_{\text{rcnt}}$ is the number of recent tokens, and $T$ is the length of the compressible region. We define $\bar{A}\in\mathbb{R}^{n_{\text{rcnt}}\times T}$ as the mean attention across heads:
\begin{equation}
\label{eq:barA}
\bar{A}_{q,t} = \frac{1}{H}\sum_{h=1}^{H} A[h,q,t],
\end{equation}
where $A[h,q,t]$ represents the attention weight from query position $q$ to key position $t$ in head $h$. Based on this, conventional KV cache eviction methods\cite{snapkv, pyramidkv} adopt the window-based attention score $s_t$ as the importance measure of token $t$:
\begin{equation}
\label{eq:st}
s_t = \sum_{q=1}^{n_{\text{rcnt}}} \bar{A}_{q,t}.
\end{equation}

In our method, the segment-level weights are modeled from two perspectives, \ie, importance and diversity. For each semantic segment $S_k$, we define:
\begin{equation}
\label{eq:Ik}
I_k = \frac{1}{|T_k|} \sum_{t\in T_k} s_t,
\end{equation}
\begin{equation}
\label{eq:Dk}
D_k = -\,\frac{1}{|T_k|}\sum_{t\in T_k}\sum_{q=1}^{n_{\text{rcnt}}},
\hat A_{q,t}\,\log\!\bigl(\hat A_{q,t}+\varepsilon\bigr)
\end{equation}
where $T_k$ denotes the set of token indices within segment $S_k$, and $\hat A_{q,t} = \frac{\bar A_{q,t}}{\sum_{q=1}^{n_{\text{rcnt}}}\bar A_{q,t}}$ is the normalized distribution of attention weights for token $t$. Here, $I_k$ measures the average attention of tokens within the segment, while $D_k$ captures the dispersion of attention distribution inside the segment.
We then combine these two indicators into segment-level weights:
\begin{equation}
\label{eq:omegak}
\omega_k = I_k \cdot \bigl(1 + \eta D_k \bigr),
\end{equation}
where $\eta \ge 0$ is a hyperparameter balancing the importance $I_k$ and diversity $D_k$. This factor strengthens segments that are both important and diverse. 

Subsequently, during token scoring, we adjust each token score using its segment factor:
\begin{equation}
\label{eq:st_tilde}
\tilde{s}_t = s_t \cdot \bigl(1 + \alpha \cdot \omega_{k(t)} \bigr),
\end{equation}
where $k(t)$ denotes the segment index of token $t$, and $\alpha \ge 0$ is a hyperparameter controlling the influence of the segment factor on token scoring. A larger $\alpha$ increases segment influence, whereas a smaller $\alpha$ emphasizes individual token scores.

This mechanism does not alter the relative ordering of tokens within each segment but raises the priority of all tokens in important segments, increasing the likelihood that entire segments are preserved in the global selection. Consequently, the final selection not only retains fine-grained token-level details but also explicitly integrates segment-level weights, effectively preventing semantic fragmentation.

\begin{algorithm}[t]
\caption{\textsc{Budget-Driven Adaptive Block Size Search}}
\label{alg:sags}
\DontPrintSemicolon
\SetAlgoLined
\textbf{Input:} Token scores $\{\tilde{s}_t\}$, segments $\{S_k\}$, token budget $B$, maximum block size $g_{\max}$, semantic fidelity threshold $\tau$ \\
\textbf{Output:} Selected block sizes and corresponding token sets $\{(g_k^*, T_k^*)\}$, the final retained token set $T^*$

\vspace{0.4em}
\textcolor{purple}{\textbf{\# Global Top-$B$ selection and implicit budgets}} \\
Compute the Top-$B$ token set $T_{\text{topB}}$ via (\ref{eq:topk}) \nllabel{sags:topk}\\
\ForEach{segment $S_k$}{
    Compute the implicit budget $b_k$ via (\ref{eq:implicit_budget}) \nllabel{sags:implicit_budget}\\
}

\vspace{0.4em}
\textcolor{purple}{\textbf{\# Block size search for each segment}} \\
\ForEach{segment $S_k$}{
  $G_k \leftarrow \{g \in \mathbb{N} \mid 1 \le g \le \min(|S_k|, b_k, g_{\max})\}$ \nllabel{sags:Gk} \\
  \For{each $g$ in descending order of $G_k$}{ \nllabel{sags:descend_g}
    Partition $S_k$ into blocks $\{C_{k,m}^{(g)}\}$ \nllabel{sags:partition} \\
    Compute block score $\Phi(C_{k,m}^{(g)}) = \sum_{t \in C_{k,m}^{(g)}} \tilde{s}_t$ \nllabel{sags:block_score} \\
    $n \leftarrow 0,\; T_k(g) \leftarrow \emptyset$ \;
        \For{block $c$ in descending $\Phi$ order}{
        \If{$n + |c| \le b_k$}{
            $T_k(g) \leftarrow T_k(g) \cup c$, \quad $n \leftarrow n + |c|$
        }
        \Else{
            Select Top-$(b_k - n)$ tokens from $c$ by $\tilde{s}_t$; add to $T_k(g)$ \nllabel{sags:block_adjust} \\
            \textbf{break}
        }
    }
    $Score_k(g) \leftarrow \sum_{t \in T_k(g)} \tilde{s}_t$ \\
    Compute token-level baseline score $Score_k^\mathrm{base} = \sum_{i \in T_{\text{topB}} \cap S_k} \tilde{s}_t$ \\
    Compute the ratio $R_k(g)$ via (\ref{eq:rk}) \nllabel{sags:Rk} \\
    \If{$R_k(g) \ge \tau$}{
      $g_k^* \leftarrow g$, \quad $T_k^* \leftarrow T_k(g)$; \\
      \textbf{break} \nllabel{sags:accept_rule}
    }
  }
}

\vspace{0.4em}
\textcolor{purple}{\textbf{\# Aggregate selected tokens}} \\
Obtain the final retained token set $T^*$ via (\ref{eq:final_selection})
\nllabel{sags:aggregate} \\
\Return $\{(g_k^*, T_k^*)\},\; T^*$
\end{algorithm}

\subsection{Budget-Driven Adaptive Block Size Search}

Building upon the enhanced token scores $\tilde{s}_t$, SABlock proposes a budget-driven adaptive block size search mechanism.
The goal is to dynamically determine the most suitable block size for each segment introduced in Section~\ref{sec:segmentation}.
We observe that larger compression budgets allow the use of larger block sizes, while tighter budgets naturally require smaller ones.
To achieve this, we design an implicit alignment strategy in two steps.
First, the number of tokens to be retained in each segment, referred to as the implicit budget, is determined through global selection.
Second, within each segment, we search for the largest block size that can effectively cover the assigned token budget.
This ensures semantic coherence while naturally aligning block size selection with budget allocation.
The procedure is summarized in Algorithm~\ref{alg:sags}.

Given a total budget $B$, we first select the Top-$B$ tokens across the entire sequence (\linerefrange{sags:topk}{sags:implicit_budget}):
\begin{equation}
\label{eq:topk}
T_\mathrm{topB}=\mathrm{TopB}_t(\tilde{s}_t), \quad |T_\mathrm{topB}|=B.
\end{equation}

Segment $S_k$ is implicitly assigned a budget, defined as the number of its tokens selected in the Top-$B$ results:
\begin{equation}
\label{eq:implicit_budget}
b_k = |T_{\mathrm{topB}} \cap S_k|.
\end{equation}

With ${b_k}$ determined, the block size ${g_k}$ is searched in each segment.
Larger $g_k$ correspond to larger blocks, whereas smaller $g_k$ correspond to finer selection, reducing to token-level methods in the extreme case.
The candidate set is defined as
$G_k = \{ g \in \mathbb{N} \mid 1 \leq g \leq \min(|S_k|, b_k, g_{\max}) \}$,
which ensures feasible block size choices for each segment while reducing the search space.

The search starts from the largest candidate block size and proceeds downward.
For each block size $g$, segment $S_k$ is partitioned into non-overlapping blocks $\{C_{k,m}(g)\}$, each scored as
$\Phi(C_{k,m}^{(g)}) = \sum_{t \in C_{k,m}^{(g)}}\tilde{s}_t$ (\lineref{sags:block_score}).
We select the highest-scoring blocks until $b_k$ tokens are covered.
If the final block exceeds $b_k$, it is refined by selecting the top tokens inside the block to precisely match the budget (\lineref{sags:block_adjust}).

To ensure quality, SABlock compares the score of the candidate set with the token-level baseline:
\begin{equation}
\label{eq:rk}
R_k(g) = \frac{Score_k(g)}{Score_k^\mathrm{base}},
\end{equation}
where $Score_k(g)=\sum_{t \in T_k(g)} \tilde{s}_t$ and $Score_k^\mathrm{base}=\sum_{t \in T_{\text{topB}} \cap S_k}\tilde{s}_t$.
Since the baseline always gives the maximum score, $R_k(g)$ acts as a fidelity measure.
A high $R_k(g)$ indicates that the compressed blocks preserve most of the important tokens, thereby ensuring semantic fidelity at larger block sizes.
If the ratio exceeds the predefined semantic fidelity threshold (i.e., $R_k(g) \ge \tau$), the block size $g$ is considered valid for segment $S_k$ and selected as a feasible choice.

The final decision for segment $S_k$ is:
\begin{equation}
\label{eq:gk_star}
g_k^* = \max \{ g \in G_k \mid R_k(g) \geq \tau \}, \quad T_k^* = T_k(g_k^*).
\end{equation}
If no candidate block size satisfies the condition, we fall back step by step to a smaller $g$ until reaching token-level selection.

Finally, the retained tokens across all segments are aggregated into a unified set (\lineref{sags:aggregate}):
\begin{equation}
\label{eq:final_selection}
T^* = \bigcup_k T_k^*, \quad |T^*| = B.
\end{equation}
These retained tokens are combined with the most recent ones to form the compressed KV cache.
In Section~\ref{ablation}, we empirically validate the effectiveness of this strategy and show that it maintains a positive correlation between budget and block size.
Specifically, when the budget is ample, the method tends to favor larger block sizes.
In contrast, under more constrained budgets, it adaptively resorts to smaller block sizes approaching token-level selection.

In summary, SABlock unifies semantic segmentation, segment-guided scoring, and adaptive block size selection into a cohesive framework.
This design allows the compression process to naturally adapt to varying memory budgets while preserving important contextual information.






\section{Performance Evaluation}
\begin{figure*}[t]
    \centering
    \scriptsize
    \setlength{\tabcolsep}{4pt}
    \renewcommand{\arraystretch}{1.2}
    \captionof{table}{Detailed accuracy comparison on 16 LongBench tasks with different KV cache budgets. 
    The optimal result is shown in \textbf{bold}, and the sub-optimal result is \underline{underlined}.}
    \label{tab:longbench_results}
    \resizebox{0.97\textwidth}{!}{%
        \begin{tabular}{l ccc ccc ccc ccc cc cc c}
        \toprule
        \multirow{2}{*}{Method} &
        \multicolumn{3}{c}{Single-Document QA} &
        \multicolumn{3}{c}{Multi-Document QA} &
        \multicolumn{3}{c}{Summarization} &
        \multicolumn{3}{c}{Few-shot Learning} &
        \multicolumn{2}{c}{Synthetic} &
        \multicolumn{2}{c}{Code} &
        \multirow{2}{*}{Avg.} \\
        \cmidrule(lr){2-4}\cmidrule(lr){5-7}\cmidrule(lr){8-10}
        \cmidrule(lr){11-13}\cmidrule(lr){14-15}\cmidrule(lr){16-17}
        & \rotcol{NrtQA} & \rotcol{Qasper} & \rotcol{MF-en}
        & \rotcol{HotpotQA} & \rotcol{2WikiMQA} & \rotcol{Musique}
        & \rotcol{GovReport} & \rotcol{QMSum} & \rotcol{MultiNews}
        & \rotcol{TREC} & \rotcol{TriviaQA} & \rotcol{SAMSum}
        & \rotcol{PCount} & \rotcol{PR-en}
        & \rotcol{Lcc} & \rotcol{R\&P}
        & \\
        \midrule
        \rowcolor{blue!8}
        \multicolumn{18}{c}{\textbf{Llama-3.1-8B-Instruct, KV cache budget=full}} \\
        \midrule
        FullKV & 21.0 & 12.78 & 27.02 & 15.83 & 16.67 & 10.02 & 34.2 & 22.59 & 26.9 & 69.5 & 91.37 & 43.89 & 6.94 & 72.79 & 65.11 & 57.14 & 37.11 \\
        \midrule
        \rowcolor{blue!8}
        \multicolumn{18}{c}{\textbf{Llama-3.1-8B-Instruct, KV cache budget=128}} \\
        \midrule
        StreamingLLM & 12.47 & 5.38 & 15.11 & 9.71 & 12.40 & 5.67 & 18.06 & 19.61 & 18.70 & 41.00 & 82.55 & 38.76 & \textbf{9.38} & 71.47 & \underline{59.06} & 48.89 & 29.33 \\
        H\textsubscript{2}O & \textbf{19.63} & 8.39 & 18.50 & 13.08 & 13.48 & 6.87 & \textbf{26.68} & 21.71 & \textbf{25.05} & 62.00 & 88.67 & 37.74 & 6.44 & 68.26 & 53.63 & 43.00 & 32.07 \\
        SnapKV & 17.28 & \underline{8.69} & 21.84 & 14.13 & 14.07 & 6.76 & 22.56 & \underline{22.54} & 21.72 & 61.00 & 90.51 & 40.29 & \underline{7.14} & 67.87 & \textbf{60.65} & \underline{49.23} & 32.89 \\
        PyramidKV & 14.15 & 8.30 & 21.65 & 12.53 & 12.96 & 7.07 & 22.39 & 22.02 & 21.65 & 63.00 & 87.44 & 40.65 & 7.00 & \textbf{71.75} & 57.23 & 47.73 & 32.41 \\
        ChunkKV & 18.05 & \textbf{8.86} & \textbf{22.39} & 12.50 & 13.42 & \underline{7.11} & 22.63 & 22.36 & 21.65 & \textbf{64.00} & 90.14 & \textbf{40.96} & 7.09 & 68.34 & 57.59 & \textbf{50.31} & \underline{32.96} \\
        SentenceKV & \underline{19.10} & 7.28 & 21.74 & \textbf{15.18} & \textbf{14.49} & \textbf{7.91} & 22.05 & \textbf{22.62} & 21.44 & 46.50 & \textbf{91.09} & \underline{40.87} & 6.48 & \underline{71.49} & 58.87 & 48.77 & 32.30 \\
        SABlock & 17.38 & 8.26 & \underline{22.04} & \underline{14.16} & \underline{14.08} & 7.10 & \underline{23.02} & 22.43 & \underline{22.18} & \underline{63.50} & \underline{90.69} & 40.82 & 7.00 & 70.69 & 58.14 & 48.17 & \textbf{33.10} \\
        \midrule
        \rowcolor{blue!8}
        \multicolumn{18}{c}{\textbf{Llama-3.1-8B-Instruct, KV cache budget=2,048}} \\
        \midrule
        StreamingLLM & 15.94 & 10.26 & 17.56 & 11.51 & 14.07 & 6.47 & 28.90 & 20.56 & 26.63 & 66.00 & 89.98 & 43.27 & \textbf{7.92} & 68.72 & 65.07 & 56.05 & 34.37 \\
        H\textsubscript{2}O & \textbf{21.44} & 11.56 & 25.23 & 15.05 & 15.47 & 8.86 & \textbf{32.09} & 23.05 & \textbf{26.96} & 69.00 & 91.53 & \textbf{44.18} & 5.21 & 70.96 & 65.22 & 56.15 & 36.37 \\
        SnapKV & 20.51 & 12.13 & \underline{26.67} & 16.44 & 15.82 & \underline{9.87} & 30.93 & 22.98 & 26.50 & 68.50 & 91.18 & 43.46 & 6.74 & 72.30 & 64.91 & 56.58 & 36.60 \\
        PyramidKV & 20.08 & \textbf{12.29} & 26.08 & 15.99 & \textbf{16.29} & 9.49 & 31.33 & \textbf{24.06} & 26.53 & 69.00 & 91.37 & 43.38 & \underline{7.06} & 72.22 & \underline{65.28} & 55.83 & 36.63 \\
        ChunkKV & \underline{20.94} & 12.04 & 26.56 & \textbf{17.21} & 15.52 & 9.59 & 31.27 & 22.95 & 26.78 & \underline{69.50} & \underline{91.54} & 43.49 & 6.44 & \textbf{72.96} & 64.84 & 56.51 & \underline{36.76} \\
        SentenceKV & 20.19 & 12.27 & \textbf{27.04} & 17.12 & 15.96 & 9.85 & 30.55 & 23.04 & 26.69 & 67.00 & \underline{91.54} & \underline{43.87} & 6.83 & \textbf{72.96} & 65.00 & \textbf{57.04} & 36.68 \\
        SABlock & 20.74 & \textbf{12.29} & 26.33 & \underline{16.72} & \underline{16.11} & \textbf{10.11} & \underline{31.66} & \underline{23.97} & \textbf{26.96} & \textbf{70.00} & \textbf{91.56} & 43.75 & \underline{7.06} & 72.22 & \textbf{65.30} & \underline{56.81} & \textbf{36.97} \\
        \midrule
        \rowcolor{yellow!15}
        \multicolumn{18}{c}{\textbf{Mistral-7B-Instruct-v0.3, KV cache budget=full}} \\
        \midrule
        FullKV & 29.06 & 41.58 & 52.88 & 49.37 & 39.01 & 28.58 & 34.97 & 25.66 & 27.78 & 76.0 & 88.59 & 47.48 & 5.0 & 98.5 & 61.48 & 62.53 & 48.03 \\
        \midrule
        \rowcolor{yellow!15}
        \multicolumn{18}{c}{\textbf{Mistral-7B-Instruct-v0.3, KV cache budget=128}} \\
        \midrule
        StreamingLLM & 21.42 & 22.18 & 26.73 & 37.73 & 33.31 & 17.61 & 16.78 & 19.79 & 17.91 & 45.50 & 85.64 & 40.52 & 5.50 & 80.00 & 54.98 & 52.02 & 36.10 \\
        H\textsubscript{2}O & 26.01 & 29.37 & 45.30 & 45.65 & 32.91 & 24.16 & 21.98 & \textbf{23.16} & 21.19 & 34.00 & 88.00 & \underline{44.17} & 6.00 & \underline{94.50} & 53.89 & 52.40 & 40.17 \\
        SnapKV & 26.78 & \underline{30.63} & 48.03 & 47.58 & 35.09 & \underline{25.35} & 21.93 & 22.12 & \textbf{21.84} & \underline{69.50} & 88.59 & 43.91 & 6.00 & 94.00 & \underline{55.70} & \underline{55.14} & 43.26 \\
        PyramidKV & \underline{26.82} & 29.77 & 48.18 & \underline{48.29} & \textbf{36.35} & 24.61 & \underline{22.31} & 21.97 & 21.33 & 69.00 & 88.80 & 44.07 & 4.50 & 94.00 & 55.39 & 52.34 & 42.98 \\
        ChunkKV & 25.83 & 29.74 & \textbf{50.69} & 46.67 & 36.08 & 25.24 & 21.89 & 22.06 & 21.21 & \textbf{71.00} & 88.08 & 43.27 & 4.00 & 93.00 & \textbf{57.57} & \textbf{57.36} & \underline{43.36} \\
        SentenceKV & 25.73 & 29.55 & 48.18 & \textbf{48.97} & 36.18 & 25.19 & 21.90 & 22.44 & 20.76 & 47.50 & \textbf{90.16} & 43.33 & \textbf{8.00} & 94.00 & 54.54 & 53.35 & 41.86 \\
        SABlock & \textbf{27.08} & \textbf{30.64} & \underline{50.19} & 47.33 & \underline{36.34} & \textbf{25.79} & \textbf{22.38} & \underline{22.68} & \underline{21.81} & 69.00 & \underline{89.02} & \textbf{44.43} & \underline{6.50} & \textbf{95.50} & 55.47 & 53.72 & \textbf{43.62} \\
        \midrule
        \rowcolor{yellow!15}
        \multicolumn{18}{c}{\textbf{Mistral-7B-Instruct-v0.3, KV cache budget=2,048}} \\
        \midrule
        StreamingLLM & 25.91 & 32.00 & 35.77 & 44.90 & 35.05 & 20.18 & 27.63 & 21.76 & 26.94 & 70.50 & 89.14 & 45.93 & 4.50 & 81.00 & \textbf{62.24} & 60.68 & 42.76 \\
        H\textsubscript{2}O & 29.60 & 38.75 & 52.79 & 49.43 & 38.39 & 27.62 & 29.02 & 24.98 & 27.59 & 63.50 & 89.11 & 47.16 & \textbf{5.50} & \textbf{99.00} & 61.46 & 61.24 & 46.57 \\
        SnapKV & \textbf{30.27} & \underline{40.52} & \underline{53.07} & 49.72 & 38.55 & 28.40 & \underline{30.91} & 25.41 & 27.38 & 75.50 & 88.86 & \textbf{47.27} & 5.00 & \textbf{99.00} & \underline{62.21} & \underline{62.23} & \underline{47.77} \\
        PyramidKV & 29.41 & 40.31 & 52.81 & 50.14 & \underline{39.57} & 28.63 & 30.22 & 25.44 & 27.44 & 75.50 & \underline{89.27} & 46.14 & 4.00 & 98.50 & 61.76 & 61.98 & 47.57 \\
        ChunkKV & 29.32 & 39.38 & 52.34 & \underline{50.19} & 38.66 & \underline{28.65} & \textbf{31.01} & \underline{25.48} & \textbf{27.84} & \textbf{76.00} & 88.64 & 47.11 & \textbf{5.50} & \textbf{99.00} & 62.07 & 62.13 & 47.71 \\
        SentenceKV & 29.72 & 40.42 & 52.75 & \underline{50.19} & \textbf{39.63} & 28.44 & 30.00 & 24.82 & 27.57 & 75.50 & 88.81 & 46.88 & \textbf{5.50} & 98.50 & 61.91 & \textbf{62.72} & 47.71 \\
        SABlock & \underline{30.26} & \textbf{41.22} & \textbf{53.38} & \textbf{50.49} & 39.36 & \textbf{29.15} & 30.62 & \textbf{25.53} & \textbf{27.84} & \textbf{76.00} & \textbf{89.57} & \textbf{47.27} & \textbf{5.50} & 98.50 & 62.11 & 62.07 & \textbf{48.01} \\
        \bottomrule
        \end{tabular}%
    }
    
    \vspace{1.5em} 
    
    \includegraphics[width=0.46\textwidth]{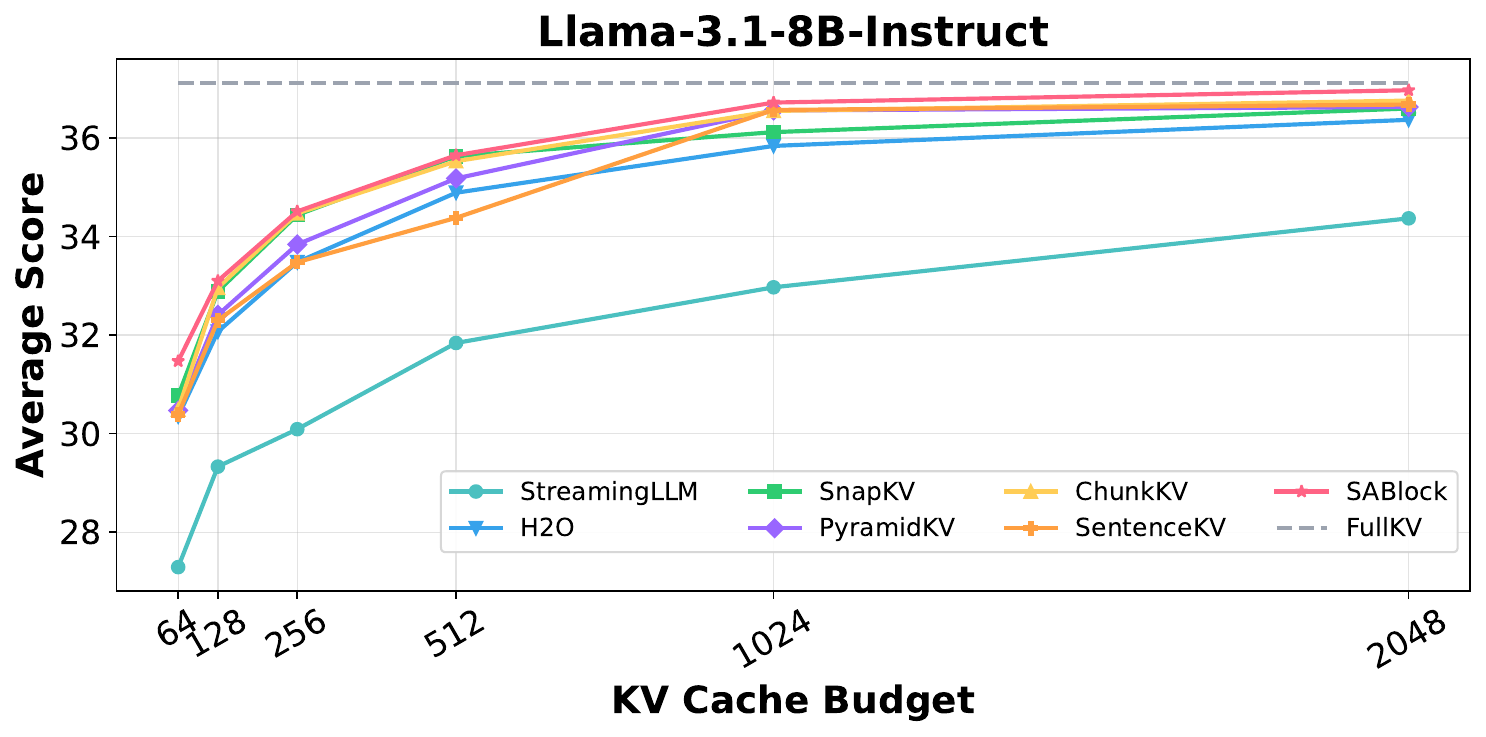}
    \hspace{0.02\textwidth} 
    \includegraphics[width=0.46\textwidth]{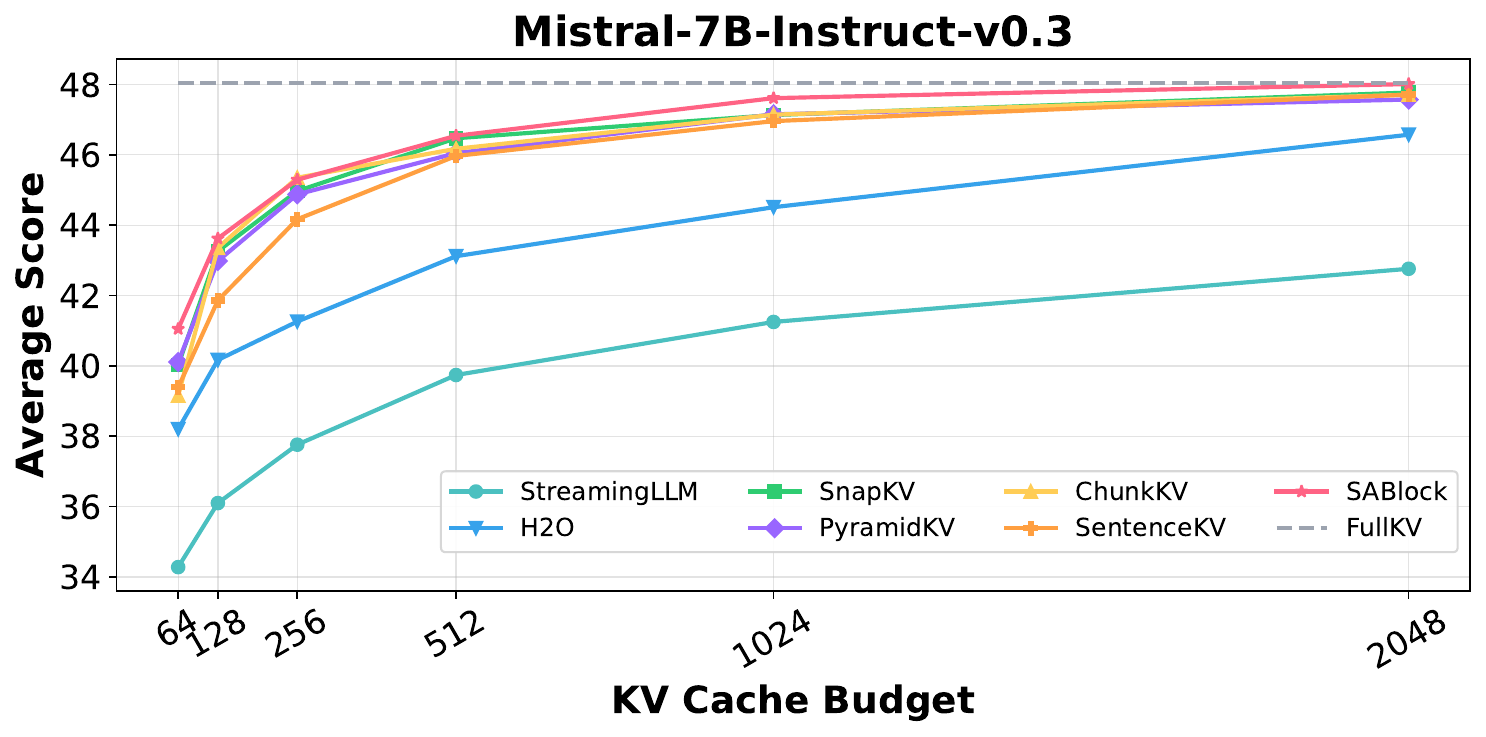}
    \captionof{figure}{Average score performance across 16 LongBench tasks under varying KV cache budgets.}
    \label{fig:longbench_overall}
\end{figure*}

\subsection{Experimental Settings}

\textbf{Experimental Models:} \ We conduct experiments on two widely adopted open-source large language models, namely \textit{Llama-3.1-8B-Instruct} \cite{dubey2024llama3} and \textit{Mistral-7B-Instruct-v0.3} \cite{jiang2023mistral}, to validate the effectiveness of our method and baselines. Both models adopt grouped-query attention (GQA) to improve decoding efficiency, but differ in their overall design (\ie, Llama uses rotary embeddings; Mistral features sliding-window attention), offering complementary perspectives on KV cache behavior across model families.

\textbf{Baseline Methods:} \ We compare SABlock with six representative methods.
(1) \textit{Position-based:} StreamingLLM~\cite{streamingllm} adopts a sliding-window strategy that retains the initial and most recent tokens.  
(2) \textit{Attention-based:} These methods score tokens with attention distributions and can be divided into three categories.
\textit{\textcircled{1} Token-level:} H\textsubscript{2}O \cite{h2o} evicts tokens by cumulative attention scores; SnapKV \cite{snapkv} scores tokens using attention scores obtained from an observation window; PyramidKV \cite{pyramidkv} assigns different cache budgets to each layer.
\textit{\textcircled{2} Block-level:} ChunkKV \cite{liu2025chunkkv} groups consecutive tokens into fixed-size blocks. 
\textit{\textcircled{3} Sentence-level:} SentenceKV \cite{zhu2025sentencekv} preserves entire sentences based on syntactic segmentation.

\textbf{Benchmark Datasets:} \ We evaluate on two representative long-context benchmarks.
(1) \textit{LongBench} \cite{bai2023longbench} tests a model's ability to reason over extended contexts across 16 tasks, spanning QA, summarization, code completion, and synthetic reasoning.
(2) \textit{Needle-in-a-Haystack (NIAH)} \cite{kamradt2023needle} focuses on retrieval and localization of critical information, assessing performance on single-needle, multi-needle, and multi-needle reasoning tasks.
These two benchmarks jointly evaluate semantic comprehension and factual retention, providing a holistic view of KV cache compression performance.

\textbf{Performance Metrics:} \ 
We report four metrics to comprehensively evaluate each method.
(1) \textit{LongBench average score}: the mean accuracy across all 16 tasks, following the official LongBench evaluation protocol \cite{bai2023longbench}, which adopts F1 for QA, Rouge-L for summarization, accuracy for synthetic reasoning, and edit similarity for code generation.
(2) \textit{NIAH accuracy}: the percentage of correctly retrieved or reasoned needles across different NIAH sub-tasks \cite{kamradt2023needle}.
(3) \textit{Peak memory usage}: the GPU memory consumed during decoding.
(4) \textit{Decoding latency}: Token-by-Token (TBT) latency under sequential decoding without batching.
\begin{figure*}[t]
  \centering
  \captionsetup[subfigure]{labelformat=empty}

  \begin{subfigure}{0.49\textwidth}
    \includegraphics[width=\linewidth]{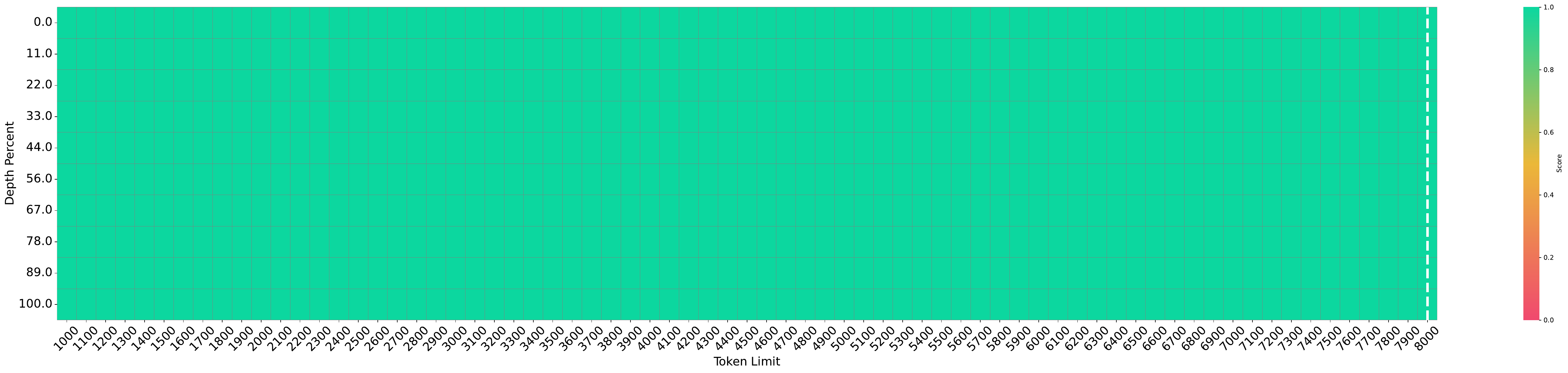}
    \caption{(a) FullKV, accuracy 100.0\%.}
  \end{subfigure}\hfill
  \begin{subfigure}{0.49\textwidth}
    \includegraphics[width=\linewidth]{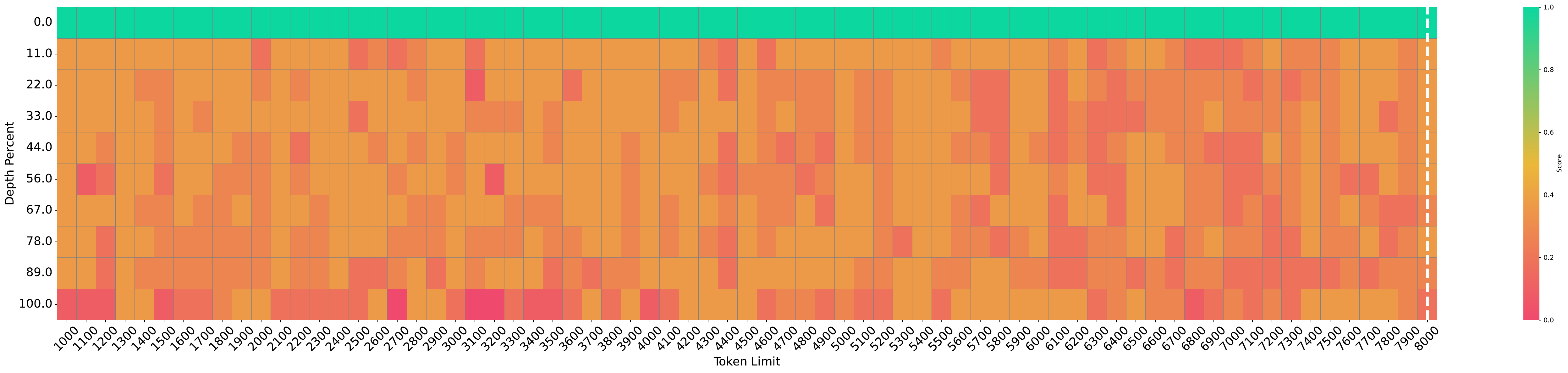}
    \caption{(b) StreamingLLM, accuracy 36.9\%.}
  \end{subfigure}

  \medskip
  \begin{subfigure}{0.49\textwidth}
    \includegraphics[width=\linewidth]{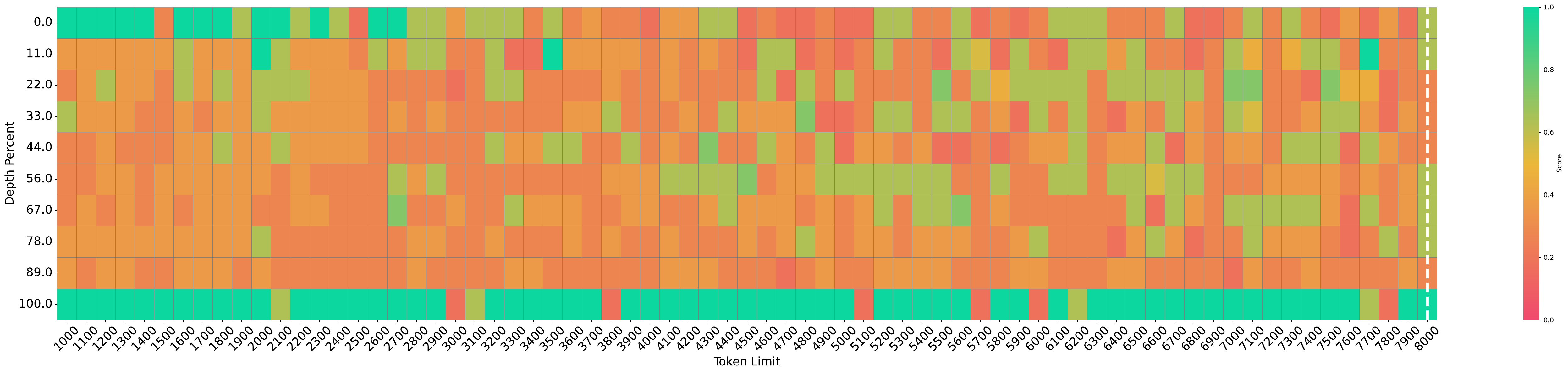}
    \caption{(c) H\textsubscript{2}O, accuracy 44.6\%.}
  \end{subfigure}\hfill
  \begin{subfigure}{0.49\textwidth}
    \includegraphics[width=\linewidth]{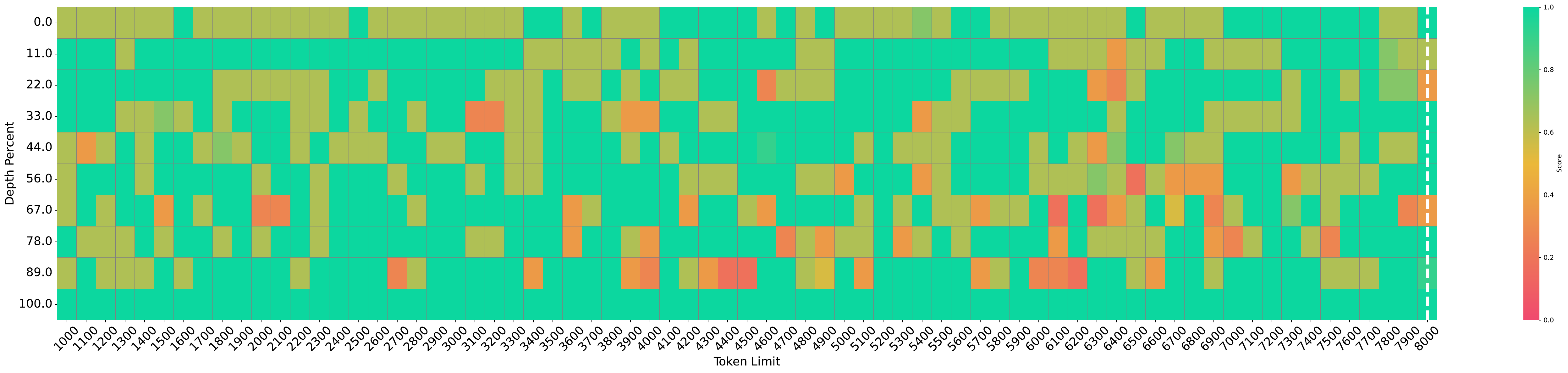}
    \caption{(d) ChunkKV, accuracy 83.3\%.}
  \end{subfigure}

  \medskip
  \begin{subfigure}{0.49\textwidth}
    \includegraphics[width=\linewidth]{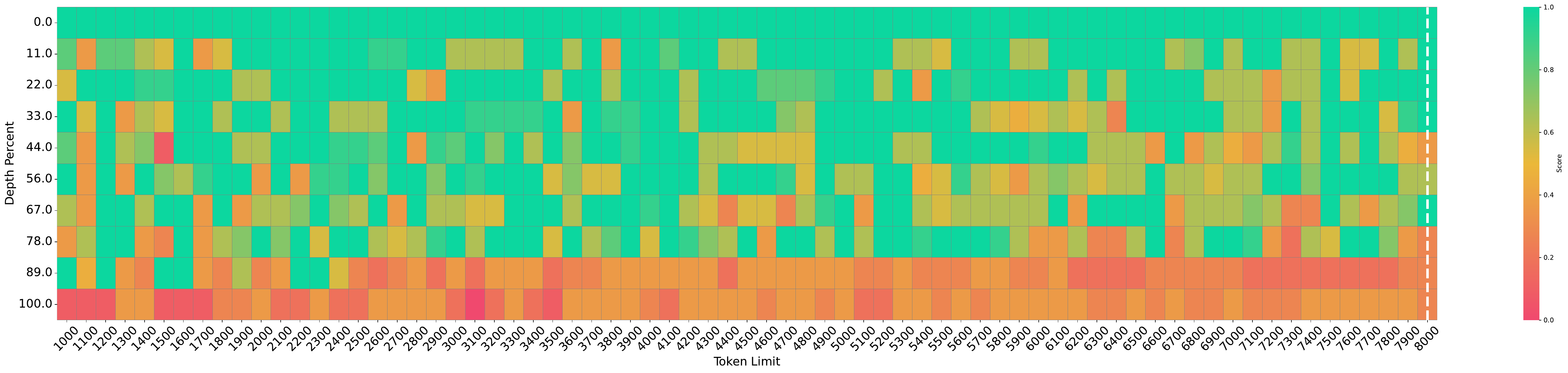}
    \caption{(e) SentenceKV, accuracy 72.1\%.}
  \end{subfigure}\hfill
  \begin{subfigure}{0.49\textwidth}
    \includegraphics[width=\linewidth]{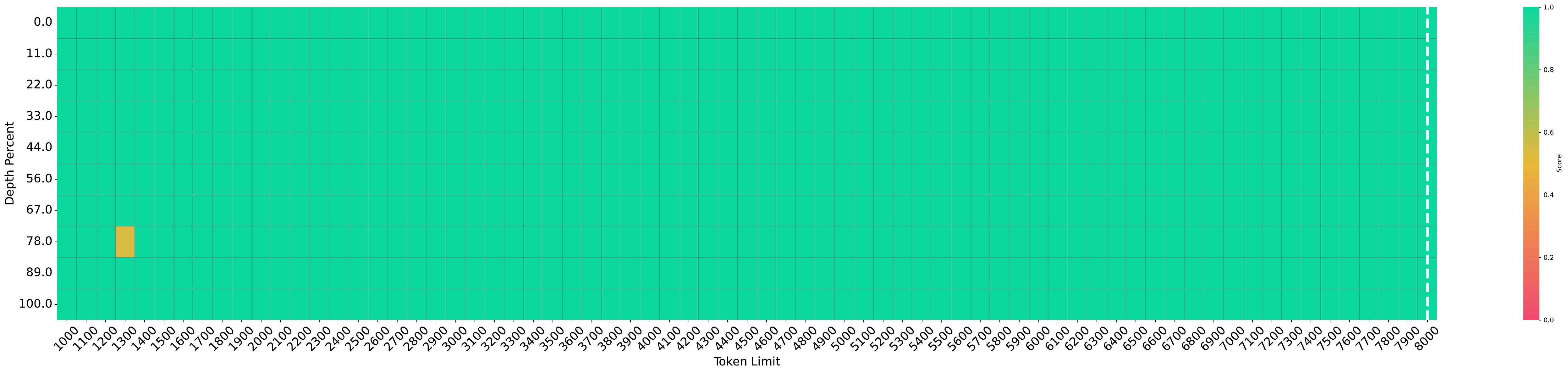}
    \caption{(f) SABlock, accuracy 99.9\%.}
  \end{subfigure}

  \caption{NIAH retrieval heatmaps on Llama-3.1-8B-Instruct with KV cache budget of 96 under 8K context length. 
  The horizontal axis denotes the total context length (in tokens), while the vertical axis denotes the relative depth at which the needle is inserted. 
  Color intensity represents the retrieval score, with greener cells indicating higher ROUGE-based similarity to the hidden needle.}
  \label{fig:niah_overall}
\end{figure*}

\textbf{Implementation Details:} \ (1) \textit{Hyperparameters Settings:} In segment-guided token scoring, we set the smoothing factor $\varepsilon=10^{-6}$ to avoid numerical instability in the entropy computation. We use $\alpha = 0.9$ to control the influence of segment-level importance, and set $\eta = 1.0$ to scale the contribution of dispersion in the segment weight calculation. In budget-driven adaptive block size search, we use a maximum block size $g_{\max}=13$, and a semantic fidelity threshold $\tau=0.85$. All these hyperparameters are determined empirically based on experimental results.
Moreover, we retain the latest $n_{\text{rcnt}}=8$ tokens of each sequence to ensure prompt stability and conversational coherence\cite{pyramidkv}. (2) \textit{Runtime Environment:} All experiments are conducted on a server with 4× NVIDIA A100 80GB PCIe GPUs and 2× Intel Xeon Platinum 8360Y CPUs \cite{liu2025adaptive}. 
We use Ubuntu 22.04, CUDA 12.6, and PyTorch 2.6.0. FlashAttention-2 is enabled for efficient long-context attention computation, and KV cache compression is applied after the prefill stage to control peak memory usage.

\subsection{Overall Performance on LongBench}

The overall evaluation results on the LongBench benchmark are presented in Table~\ref{tab:longbench_results} and Fig.~\ref{fig:longbench_overall}. The table reports detailed scores across different tasks, while the figure summarizes the average score performance over 16 sub-tasks under varying cache budgets.
In general, the performance of all methods improves as the cache budget increases, but the degree of improvement and relative differences vary significantly. 
Traditional approaches such as StreamingLLM and H\textsubscript{2}O suffer from a sharp accuracy drop under extremely small cache budgets (\eg, 128), as they can only retain partial local information and fail to preserve long-range dependencies. 
SnapKV and PyramidKV exhibit more stable performance with larger caches, but their reliance on fixed block sizes imposes bottlenecks in tasks such as multi-document QA and summarization. 

In contrast, SABlock maintains high performance even under tight cache budgets and gradually approaches the full-cache baseline as the cache budget increases. For example, with only 2,048 KV entries preserved, it significantly reduces memory consumption while still matching or outperforming FullKV. 
Moreover, with 1.8\% of the KV cache retained (KV cache budget = 128), the performance drops by only 9.99\% compared to the full-cache baseline.
In summary, SABlock consistently demonstrates a leading trend across different model architectures, indicating that its semantic-aware adaptive block size strategy effectively avoids semantic fragmentation and ensures good generalization.

\subsection{Context Retrieval under NIAH}

We further evaluate SABlock on the NIAH benchmark to assess its retrieval capability under long-context settings. The NIAH test is designed to examine a model’s ability to accurately retrieve embedded key information from lengthy documents, where the challenge lies in the random distribution and variable depth of the target information. This poses stricter requirements on the information retention ability of KV cache compression methods.
In this experiment, we conduct evaluations using the Llama-3.1-8B-Instruct model with a context length of 8K tokens and a fixed KV cache budget of 96, comparing against baselines including FullKV, StreamingLLM, H\textsubscript{2}O, ChunkKV, and SentenceKV. Even with 80GB of GPU memory, it is infeasible to concurrently process 20 inference requests with an 8K context length, highlighting the necessity of efficient KV cache compression. 

As shown in Fig.~\ref{fig:niah_overall}, SABlock achieves an accuracy of 99.9\% with only 96 KV entries, closely matching FullKV (100\%) and significantly outperforming other baselines such as StreamingLLM (36.9\%) and H\textsubscript{2}O (44.6\%). SABlock consistently retrieves the target information (needle) across varying insertion depths and sequence lengths, exhibiting strong robustness. For instance, even when the needle is deeply buried in a 7K+ token document, SABlock retains sufficient KV entries for the relevant segment and reconstructs its content through coarse-grained blocks.

We attribute this strong performance to two key factors: (1) semantic segmentation ensures that needles embedded within coherent spans are preserved as a whole, reducing fragmentation; (2) adaptive block size allows the use of larger blocks when segments contain concentrated importance, increasing coverage efficiency. Compared to other approaches, SABlock dynamically allocates capacity in a content-aware manner, making it particularly well-suited for retrieval-heavy long-context scenarios.
\begin{figure*}[t]
    \centering
    \begin{minipage}{0.62\textwidth}
        \centering
        \includegraphics[width=\linewidth]{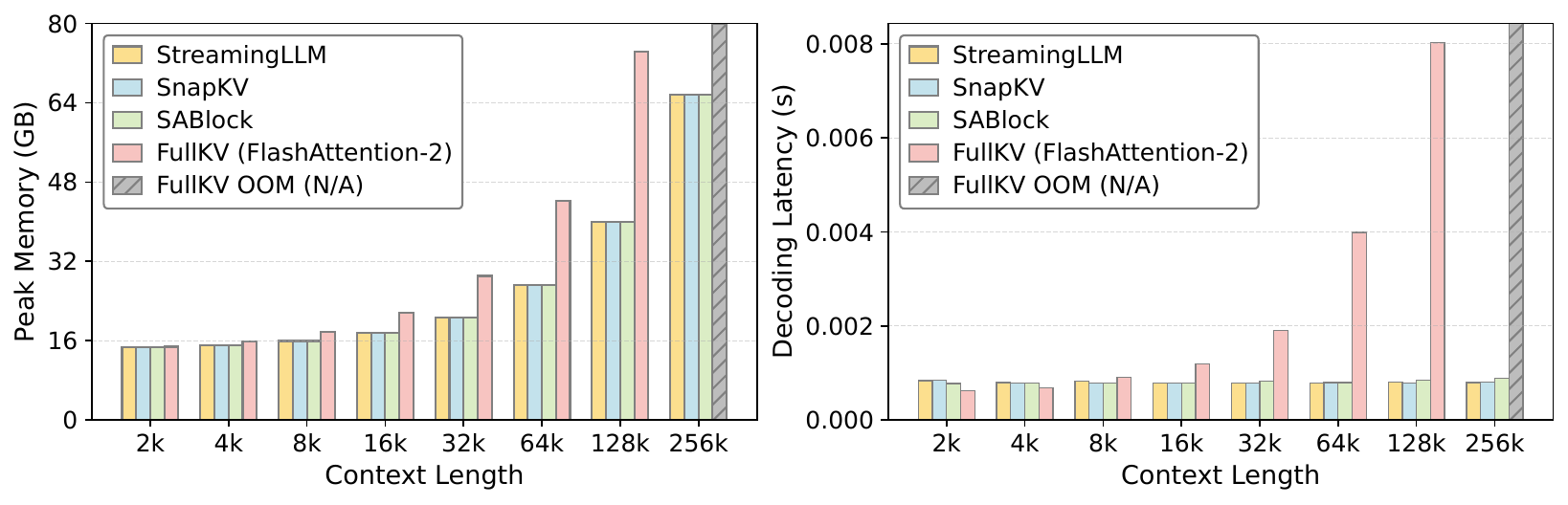}
        \caption{Peak memory usage (left) and decoding latency (right) on a single A100 GPU across different context lengths.}
        \label{fig:memory_latency}
    \end{minipage}%
    \hfill
    \begin{minipage}{0.36\textwidth}
        \centering
        \renewcommand{\arraystretch}{1.15}
        \setlength{\tabcolsep}{4pt}
        \scriptsize
        \captionof{table}{Ablation study of SABlock.}
        \label{tab:ablation}
        \begin{tabularx}{\linewidth}{X c c}
            \toprule
            \textbf{Approach} & \textbf{Avg. Score} & \textbf{Relative Drop} \\
            \midrule
            SABlock                   & 33.78 & -- \\
            w/o Segment-Guided Scoring & 33.03 & $-$2.22\% \\
            w/o Adaptive Block Size    & 33.36 & $-$1.24\% \\
            \bottomrule
        \end{tabularx}
        
        \vspace{1em} 

        \renewcommand{\arraystretch}{1.15}
        \setlength{\tabcolsep}{4pt}
        \scriptsize
        \captionof{table}{Block size ablation.}
        \label{tab:ablation2}
        \begin{tabularx}{\linewidth}{X c c}
            \toprule
            \textbf{Approach} & \textbf{Avg. Score} & \textbf{Relative Drop} \\
            \midrule
            Adaptive (Ours)             & 33.78 & -- \\
            Token-level                 & 32.15 & $-$4.83\% \\
            Block-level (block size=5)  & 32.81 & $-$2.87\% \\
            Block-level (block size=10) & 32.20 & $-$4.68\% \\
            \bottomrule
        \end{tabularx}
    \end{minipage}
    \vspace{-2mm}
\end{figure*}

\subsection{Evaluation of Resource Consumption}

To comprehensively assess the efficiency improvements brought by SABlock, we evaluate the resource consumption from two aspects: \textit{peak memory} and \textit{decoding latency}. All experiments are conducted using Mistral-7B-Instruct-v0.3 with FlashAttention-2 \cite{dao2023flashattention2}, running on a single NVIDIA A100 80GB GPU. Except for the full-cache baseline, all methods use a fixed KV cache budget of 1,024. 

\textbf{Peak Memory:} \ 
As shown in the left plot of Fig.~\ref{fig:memory_latency}, SABlock consistently reduces peak memory usage across various context lengths. In long-context scenarios (\eg, 128K tokens), SABlock significantly reduces the budget of the KV cache compared to FullKV, resulting in a 46.28\% drop in peak memory usage and effectively alleviating memory bottlenecks. This improvement is crucial in practice. For example, a 256K context length already leads to GPU memory overflow on A100, which is roughly equivalent to handling 32 inference requests with an 8K context length (batch size is 32), a highly common serving scenario. 
By compressing the KV cache, SABlock not only supports longer contexts but also enables higher concurrency under limited hardware memory, making efficient deployment feasible.

\textbf{Decoding Latency:} \ 
The right plot of Fig.~\ref{fig:memory_latency} illustrates decoding latency. As input length increases, FullKV experiences a sharp rise in latency due to attention computation costs. In contrast, SABlock compresses the KV cache after the \textit{prefill} stage, effectively reducing the computational burden per decoding step. At an input length of 128K tokens, latency is reduced by 89.49\%, achieving a 9.5$\times$ speedup. This brings tangible benefits to real-time applications such as dialogue systems, where fast token generation is crucial.

\subsection{Ablation Studies}\label{ablation}

To evaluate the contribution of each module in SABlock, we conduct ablation experiments based on Llama-3.1-8B-Instruct under a fixed KV cache budget of 96, using the LongBench average score as the metric.
SABlock is composed of two key components: (1) \textit{Segment-Guided Token Scoring}, which enhances token importance by injecting segment-level semantics, and (2) \textit{Budget-Driven Adaptive Block Size Search}, which dynamically selects the block size per segment according to retained positions and budget availability.
We first analyze the effectiveness of these two mechanisms, and then further examine the block size selection trends under varying budgets to demonstrate that our search strategy naturally adapts block size choices to the budget, which in turn helps maintain generation quality.


\textbf{Effectiveness of Segment-Guided Token Scoring:} \ 
Table~\ref{tab:ablation} shows that removing the segment-guided token scoring mechanism (\textit{w/o Segment-Guided Scoring}) leads to a performance drop of 2.22\%. 
This validates that relying solely on token-level attention scores can cause semantic fragmentation, as globally important segments may contain some tokens with relatively low scores. 
By propagating segment-level importance and diversity into token scoring, this mechanism raises the collective priority of key segments, ensuring that semantically complete chunks are preserved in global selection.

\textbf{Effectiveness of Budget-Driven Adaptive Block Size Search:} \ 
As shown in the same table, disabling adaptive block size selection (\textit{w/o Adaptive Block Size}) causes a 1.24\% drop in LongBench average score. 
To further analyze different strategies, Table~\ref{tab:ablation2} compares our adaptive approach with token-level and fixed block-level baselines ($g{=}5,10$). 
The adaptive strategy consistently outperforms these alternatives. 
While token-level compression captures fine-grained details, it risks breaking semantic continuity across tokens. 
Fixed block-level methods, although simple, misalign with semantic boundaries and require careful tuning of block size. 
Our adaptive mechanism effectively balances fidelity and efficiency by tailoring the block size to both content structure and memory constraints.

\begin{figure}[t]
    \centering
    \includegraphics[width=0.98\linewidth]{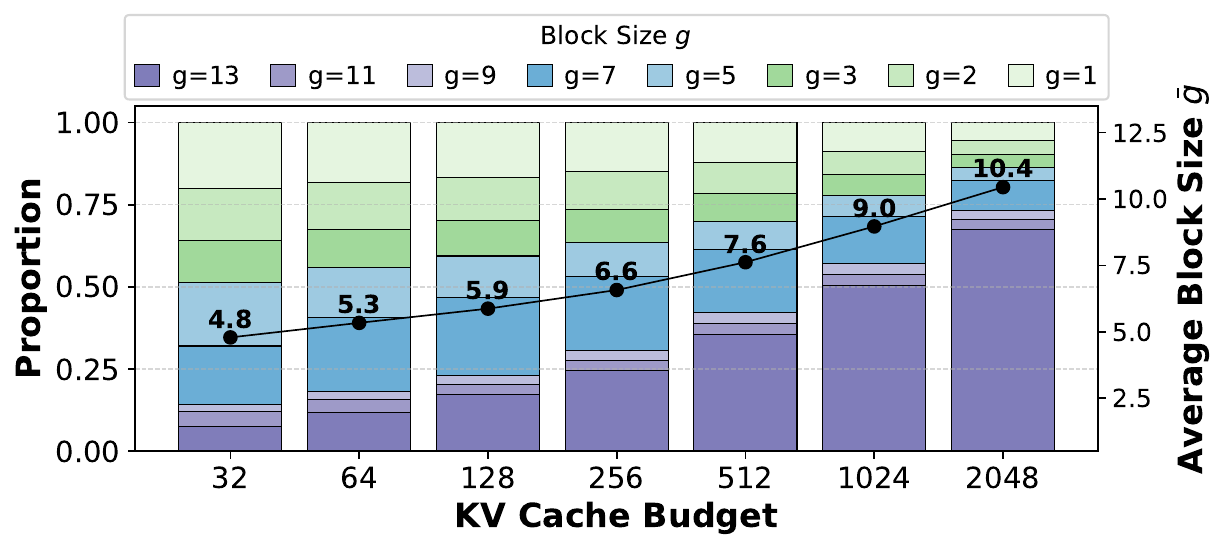}
    \caption{Stacked proportions of selected block sizes $g$ under different KV cache budgets $K$ (left $y$-axis). The solid line (right $y$-axis) shows the average block size $\bar{g}$, indicating that larger budgets favor larger blocks while smaller budgets favor smaller ones.}
    \label{fig:budget_granularity_comp}
    \vspace{-2mm}
\end{figure}
\textbf{Block Size Selection Trends:} \ 
Fig.~\ref{fig:budget_granularity_comp} illustrates the distribution of selected block sizes under varying cache budgets $K$. 
As the budget increases, larger block sizes (\eg, $g{=}13$) are used more frequently, while smaller block sizes (\eg, $g{=}1,2,3$) dominate under tighter budgets. 
This trend naturally arises from our top-down search procedure, where larger blocks are preferentially selected when they can efficiently cover global positions without violating segment-level fidelity. 
To balance expressiveness and efficiency, we limit the search space to $\{1,2,3,5,7,9,11,13\}$, which still provides sufficient coverage. 
These results demonstrate that SABlock’s compression block size adapts naturally to the available memory budget, without requiring explicit budget segmentation or cross-segment coordination.

\subsection{Sensitivity Analysis}

\begin{figure}[t]
  \centering
  \begin{subfigure}{0.48\linewidth}
    \centering
    \includegraphics[width=\linewidth]{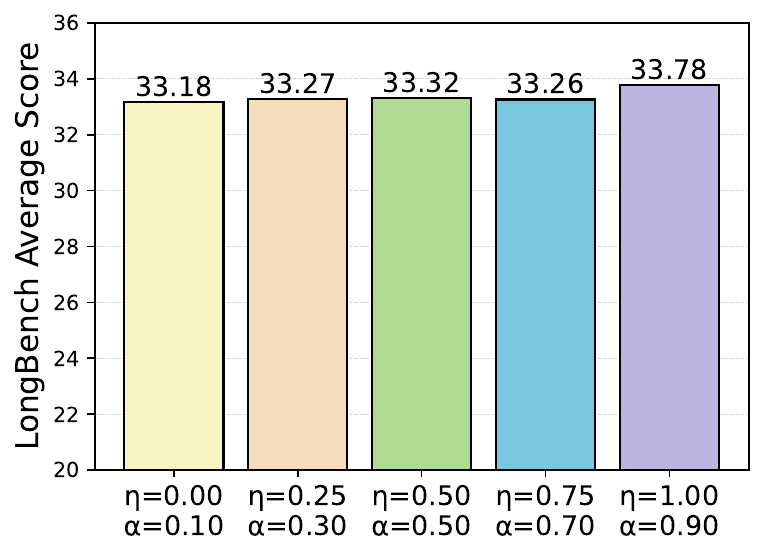}
    \caption{Impact of $\eta$ and $\alpha$.}
  \end{subfigure}
  \hfill
  \begin{subfigure}{0.48\linewidth}
    \centering
    \includegraphics[width=\linewidth]{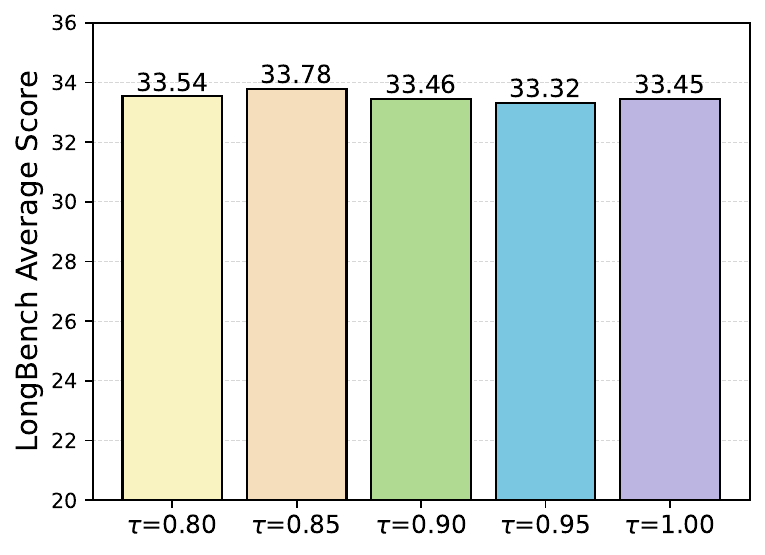}
    \caption{Impact of $\tau$.}
  \end{subfigure}
  \caption{Sensitivity of SABlock under KV budget of 96. Bars report LongBench average scores.}
  \label{fig:sensitivity}
\end{figure}

To assess the robustness of SABlock to various hyperparameter settings, we conduct sensitivity analyses on its two major components: (1) segment-guided token scoring, which involves weighting factors $\eta$ and $\alpha$, and (2) budget-driven block size search, which is controlled by the semantic fidelity threshold $\tau$. All experiments are conducted under a fixed KV cache budget of 96 using the Llama-3.1-8B-Instruct model on LongBench. As shown in Fig.~\ref{fig:sensitivity}, SABlock maintains stable performance across a wide range of values, indicating low sensitivity and ease of deployment.

\textbf{Impact of Weighting Factors $\eta$ and $\alpha$ in Segment-Guided Scoring:} \ 
Fig.~\ref{fig:sensitivity}(a) presents the LongBench average scores under various combinations of $\eta$ and $\alpha$. All tested configurations yield scores around 33 with minor fluctuations, demonstrating strong robustness of our method. Along the $\alpha$ dimension, increasing its value from 0.1 to 0.9 does not lead to significant changes. Only when $\alpha$ is too small (\eg, 0.1) does the score drop slightly, indicating that the segment-guided adjustment becomes too weak, and the behavior degenerates to token-level scoring. Along the $\eta$ dimension, which controls the influence of intra-segment attention diversity, performance remains highly stable. This suggests that once the segment-level signal is sufficiently strong, additional diversity weighting introduces little variance. 

\textbf{Impact of Semantic Fidelity Threshold $\tau$ in Block Size Search:} \ 
Fig.~\ref{fig:sensitivity}(b) reports the impact of $\tau$ on compression behavior and downstream performance.
A smaller $\tau$ leads to more aggressive compression with larger block sizes, potentially sacrificing fine-grained details;
a larger $\tau$ favors smaller block sizes and higher local fidelity but risks fragmenting semantically coherent units.
We observe that performance varies smoothly within the range $\tau \in [0.80, 1.00]$, with the best overall result achieved at $\tau \approx 0.85$, which we set as the default choice in all experiments.
This balanced value allows SABlock to retain sufficient semantic information while still enabling effective compression.

Across all tested hyperparameters, SABlock demonstrates low sensitivity and stable performance. These results validate SABlock as a practical and reliable KV cache compression solution, capable of maintaining high quality under diverse conditions and strict resource constraints.

\section{Related Work}
\textbf{Efficient Inference for LLMs:} \  LLMs face substantial computational and memory overhead during inference, which has motivated a variety of efficient inference techniques. To address the high complexity of attention, existing approaches propose variants such as Multi-Query Attention~\cite{ainslie2023gqa, shazeer2019fast}, Kernel-based Attention~\cite{choromanski2020performer}, and Low-Rank Attention~\cite{wang2020linformer, ma2021luna}, which reduce computational and memory costs. Another line of work focuses on dynamic inference by adaptively selecting model sub-structures based on the input, where early exiting~\cite{fan2024not, elhoushi2024layerskip} allows models to terminate inference at different layers, thereby reducing computation and latency. At the system level, operator fusion~\cite{luo2025clusterfusion, salmani2025llm} reduces memory accesses and kernel launch overhead, scheduling and batching~\cite{kwon2023pagedattention, holmes2024deepspeedfastgen, modeltc2024lightllm} improve hardware utilization, and memory management~\cite{liu2024intactkv, wang2024model, snapkv, infllm} alleviates the long-context bottleneck by compressing the KV cache through techniques such as quantization, merging, eviction, or offloading. Our work primarily focuses on KV cache eviction within memory management and proposes a new eviction framework that reduces memory requirements during inference.

\textbf{KV Cache Eviction:} \ The core objective of KV cache eviction is to reduce memory usage while preserving the critical information required for generation. Early approaches, such as StreamingLLM~\cite{streamingllm}, adopt a sliding-window strategy that retains the initial and most recent tokens, but this ignores semantic importance and struggles with tasks requiring long-range dependencies. Later methods began to score token importance: H\textsubscript{2}O~\cite{h2o} removes low-scoring tokens based on cumulative attention scores, SnapKV~\cite{snapkv} leverages an observation window to extract attention distributions and identify key tokens, PyramidKV~\cite{pyramidkv} assigns different cache budgets across layers in a hierarchical manner to improve the preservation of critical information, and ChunkKV~\cite{liu2025chunkkv} groups consecutive tokens into fixed-size blocks to preserve local semantic continuity. However, most of these methods perform eviction at a fixed block size (either token-level or block-level), leaving adaptive block size largely unexplored. To address the semantic fragmentation caused by fixed-block-size compression, SentenceKV~\cite{zhu2025sentencekv} performs compression at the sentence level, treating each sentence as a unit. However, it still suffers from redundant retention when sentences are long. In contrast, SABlock innovatively introduces semantic-aware adaptive block size compression, improving memory efficiency while preserving contextual coherence.

\section{Conclusions}
In this work, we propose SABlock, a semantic-aware KV cache eviction framework with adaptive block size, to address the limitations of existing token-level, block-level, and sentence-level compression methods. By dynamically selecting the block size based on segment-level semantics and global memory budgets, SABlock effectively balances memory efficiency and contextual coherence. Extensive experiments on long-context benchmarks demonstrate that SABlock significantly outperforms state-of-the-art baselines on diverse benchmarks and under different memory constraints.

\bibliographystyle{IEEEtran}
\bibliography{ref}

\end{document}